\documentclass[journal]{IEEEtran}
\ifCLASSINFOpdf
  \usepackage[pdftex]{graphicx}
\else
  \usepackage[dvipdfmx]{graphicx}
\fi
\usepackage{amsmath}
\usepackage{array}

\ifCLASSOPTIONcompsoc
  \usepackage[caption=false,font=normalsize,labelfont=sf,textfont=sf]{subfig}
\else
  \usepackage[caption=false,font=footnotesize]{subfig}
\fi
\usepackage{url}

\usepackage{multirow}
\usepackage[switch]{lineno}

\begin{document}
%
\title{Behavioral assessment of a humanoid robot when attracting pedestrians in a mall}

%
%
%

\author
    {Yuki~Okafuji,
      Yasunori~Ozaki,
      Jun~Baba,
      Junya~Nakanishi,
      Kohei~Ogawa,
      Yuichiro~Yoshikawa,
      Hiroshi~Ishiguro
      \thanks{Y. Okafuji is with the School of Information Science and Engineering, Ritsumeikan University, Shiga, Japan e-mail: yokafuji@fc.ritsumei.ac.jp.}
      \thanks{Y. Ozaki and J. Baba are with AI Lab, CyberAgent, Inc., Tokyo, Japan.}
      \thanks{J. Nakanishi, Y. Yoshikawa, and H. Ishiguro are with the Graduation School of Engineering Science, Osaka University, Osaka, Japan.}
      \thanks{K. Ogawa is with the Graduation School of Engineering, Nagoya University, Aichi, Japan.}
    }

\maketitle

\begin{abstract}
Research currently being conducted on the use of robots as human labor support technology. In particular, the service industry needs to allocate more manpower, and it will be important for robots to support people. This study focuses on using a humanoid robot as a social service robot to convey information in a shopping mall, and the robot's behavioral concepts were analyzed. In order to convey the information, two processes must occur. Pedestrians must stop in front of the robot, and the robot must continue the engagement with them. For the purpose of this study, three types of autonomous behavioral concepts of the robot for the general use were analyzed and compared in these processes in the experiment: active, passive-negative, and passive-positive concepts. After interactions were attempted with 65,000+ pedestrians, this study revealed that the passive-negative concept can make pedestrians stop more and stay longer. In order to evaluate the effectiveness of the robot in a real environment, the comparative results between three behaviors and human advertisers revealed that (1) the results of the active and passive-positive concepts of the robot are comparable to those of the humans, and (2) the performance of the passive-negative concept is higher than that of all participants. These findings demonstrate that the performance of robots is comparable to that of humans in providing information tasks in a limited environment; therefore, it is expected that service robots as a labor support technology will be able to perform well in the real world.
\end{abstract}


\section{Introduction}
\label{sec1}
\IEEEPARstart{S}{ocial} robots have been developed and are presently being used in our daily lives. These social robots are deployed in the service industry for various purposes. For instance, social service robots are used as museum guides~\cite{Shiomi07}, travel guides~\cite{Triebel16}, shopping guides~\cite{Gross08}, and for hotel services~\cite{Osawa17}. Currently, significant advancements in machine learning have improved the performance of robots; however, there are limitations for the tasks that robots can perform. In the long-term perspective, however, robots will gradually improve in performance and become more necessary in our lives. As robots become more widespread in our lives, they are expected to become a labor support technology in society~\cite{Yamazaki12} and provide a new type of customer service as avatar robots~\cite{Baba20}.

Among the various tasks in the service industry, tasks of providing information and presenting advertisements in commercial facilities are expected to be one of the roles that robots can play~\cite{Kanda10}. It is expected that a robot is able to approach a target customer more efficiently by using its embodiment, as opposed to the conventional method of providing information through papers or digital signages. For example, some applications of robots wherein they are used for a variety of services (such as customers favor purchasing status goods and order and eat more food) have been validated~\cite{Mende19}. Moreover, stakeholders, which includes customers, shop managers, and mall managers, are positive about introducing robots in shopping malls~\cite{Niemela17, Niemela19}. Consequently, introducing robots into commercial facilities is not limited to advertising and providing information, but they are also highly expected to be a part of our future. Despite these expectations, it has been reported that social robots in the real world tend to be neglected by users even if the robots talk to them, owing to the limitations of their abilities~\cite{Lee12, Tanaka16}. To maximize the effectiveness of robots, it is necessary to not only improve their capabilities but also to identify the types of behavior they should display. 
\begin{figure}[!t]
  \centering
  \includegraphics[width=2.5in]{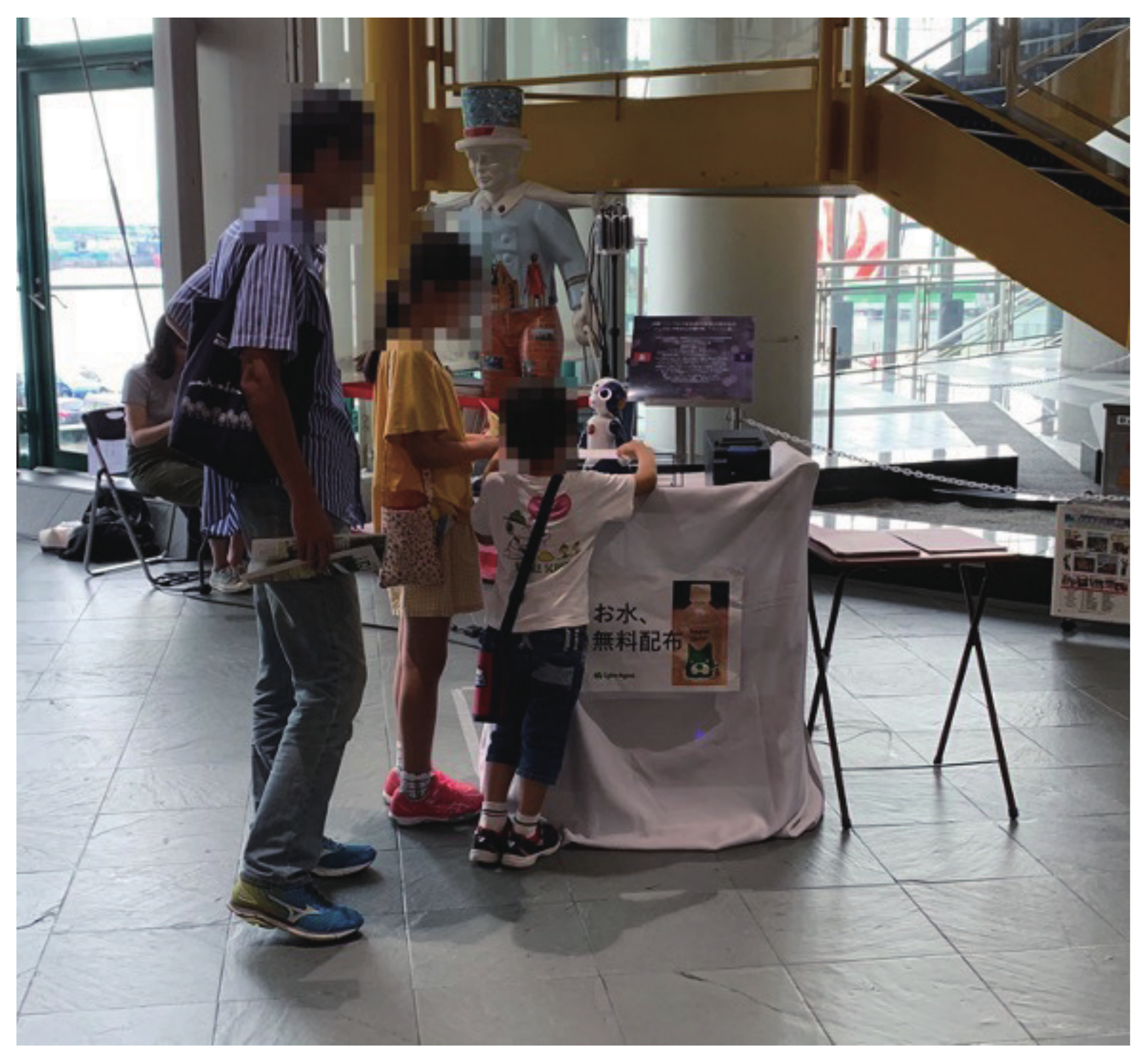}
  \caption{One of the experimental scenes in a shopping mall. The autonomous robot attempts to convey information about the mall to pedestrians.}
  \label{fig:scene}
\end{figure}

In this study, therefore, we investigated robot behaviors in terms of providing information and advertising to pedestrians in a shopping mall. The setup of the experiment is presented in Fig~\ref{fig:scene}. In particular, the detailed aims of this study are twofold:
\begin{enumerate}
  \item Explore versatile robot behaviors that do not depend on providing specific types of information or advertisements.
  \item Propose and verify the behavioral concepts of the robot rather than detailed behavioral verifications, such as eye-contact and talk timing.
\end{enumerate}
We believe that these aims will lead to a discussion on how robot behaviors can be applied in general.

In this study, there are three main steps in which robots successfully provide information. These steps are drawing the attention of pedestrians to the robot, making pedestrians stop in front of the robot, and continuing the engagement until the message is delivered. In these tasks, we believe that getting pedestrians to stop and conveying the information to the end are the most difficult tasks. By providing an opportunity for the robot to perform well for these two tasks, we proposed three types of behavioral concepts. These concepts are active, passive-negative, and passive-positive concepts, and they were verified through a field experiment. Based on these results, we discuss what kind of robot behavior concepts can trigger people to want to communicate with robots. In Section~\ref{sec3}, we perform experiments on the robots of the three types of robot behavior for providing information that have been verified in a shopping mall. We also examined whether the proposed robot system was more effective than human advertisers in terms of providing information. For robots to support human labor, they should to demonstrate performance equivalent to that of humans. In particular, tasks in which the robot has a superior performance must be identified. Therefore, four human advertisers were gathered, and the experiment was conducted under the same conditions as that of the robot. Then, we compared the robot's performance to humans. The human experiments are described in Section~\ref{sec4}. The rest of the papr is structured as follows. In Section~\ref{sec2}, related works are described. In Section~\ref{sec5}, the discussion based on the results in the two experiments is provided. Finally, Section~\ref{sec6} provides the conclusions and describes future work.

The preliminary research for this study was presented at a conference and published in the proceedings~\cite{Okafuji20} in which we reported our limited results of the pedestrians affected by the robot behaviors, shown in Table~\ref{tab:results1} and Fig.~\ref{fig:whole_robot} in Section~\ref{sec3}. The current paper provides a detailed analysis depending on the gender, age, and characteristics of the pedestrians in Section~\ref{sec3}. In addition, the current paper provides a comparison of the performance results between the robots and the humans in Section~\ref{sec4}. The comparative results in Section~\ref{sec4} are crucial for the discussion section as an insight into the competence of robots and the diffusion of robots in society. Based on the added results, the introduction, related works, discussion, and conclusion sections are also refined.

\section{Related Works}
\label{sec2}
\subsection{Service Robot in Real-World Environment}
There are many examples of robots being used to provide information and display advertisements in stores. One of the aims of this research in real-world environments is to build the robot system itself. For instance, experiments on a semi-autonomous robot are supported by human operators for dialogues. These robots have been placed at shopping malls and stations~\cite{Kanda10, Shiomi08}, and a multiple robot system has been implemented in a shopping mall~\cite{Shiomi09}. These studies aimed to verify the effectiveness of the system itself, i.e., whether the robot system can be used in the real-world environment, rather than to verify the effects of the robot's detailed behavior.

Several previous studies focused on comparing and verifying the detailed behaviors of the robots that are effectively used by users in real environments. An example of a robot using its mobility is a robot handing out flyers~\cite{Shi13, Shi18}. The behavior of humans distributing flyers was analyzed, and the analyzed optimal behavior was implemented into a robot and then tested in a shopping mall. The robots in the study applied their mobility capabilities to the maximum effect in the task of distributing flyers.

In contrast, stationary robots (or robots that rarely move) have difficulty in terms of providing information because their actions are limited because they cannot approach pedestrians themselves. These robots need to attract pedestrians and draw them near the robot through their presence and actions. After engaging with pedestrians, a study in a museum~\cite{Yamazaki08} explored the timing of the robot's head and gaze action to increase a pedestrian's engagement. In the shopping mall, robot behaviors that are natural for humans have also been validated when humans approach robots~\cite{Bergstrom09}. With respect to the behavior of the stationary robot before engaging pedestrians, the social presence of robots is important; thus, the talking behavior of a robot has been shown to be effective~\cite{Iwasaki18}. In addition, a looking-back behavior is also effective for gaining pedestrians' attention~\cite{Iwasaki19}.

A more detailed analysis shows that there are five elements of robot behaviors that are important in terms of engaging with users: eye contact, duration of eye contact, distance to users, approaching users, and laughing. Shiomi et al. investigated different robot sizes and different conversational schemes for providing information to examine detailed robot behavior~\cite{Shiomi13}. This study revealed that smaller robots that provide specific information have a high success rate in terms of delivering information. Hayashi et al. validated the effectiveness of providing information in terms of the number of robots and the conversation methods~\cite{Hayashi07}. The results suggest that a passive-social medium, where two robots talk to each other and they can provide information to pedestrians indirectly, is better than an interactive-social medium with the users. There is also a study of virtual agents that can change their behavior that is based on the spatial characteristics of humans to enable robots to engage with humans more effectively~\cite{Michalowski06}.

Previous studies of stationary robots have often focused on developing robots that can gain the attention or engagement of pedestrians by implementing and comparing detailed robot behaviors. However, although specific behaviors such as looking-back behavior tend to encourage pedestrian engagement, other robots may be unable to implement them due to functional limitations. Therefore, to implement common behaviors to more robots, it is important to propose a behavioral concept such as the passive-social medium. If we can propose an effective behavioral concept, we can build a variety of detailed motion patterns according to the characteristics of each robot that is based on the concept. In other words, the concepts can be used more generically. A few studies have compared and validated these behavioral concepts. However, it is well known that studies on human-robot interaction are highly influenced by cultural differences~\cite{Li10, Trovato13}. Therefore, additional studies across various places should be conducted in the future.

\subsection{Performance Comparison between Robots and Humans}
When developing a service robot to operate in a real environment, most research aims to be able to accomplish tasks such as providing information and distributing flyers. However, if robots are used as a labor support technology, it is important to compare the performance of robots and humans, and a few studies aimed at them.

As one of the studies that directly compared task performance, an android robot as a salesperson attempted to sell goods at a department store~\cite{Watanabe15}. This study shows that the android robot is able to sell goods as well as human selling goods performance. Another study compared a teleoperated robot with a human in the task of distributing food samples~\cite{Tonkin17}. By distributing food samples while the robot passively approaches pedestrians, the robot achieves high performance comparable to humans. These differences in performance have been shown to be influenced by the unique characteristics of the robot. Factors such as the eeriness of the robot can cause discomfort to the consumer, and as a result, compensatory consumer response such as ordering more food is facilitated~\cite{Mende19}. In other words, the high task performance of robots is not only due to their high ability, but also due to a variety of other factors.

There are also some studies that have investigated how users feel when robots (and virtual agent) and humans perform the same task, although they are not directly compared in task performance. A comparison of tasks performed by a virtual agent and a human in service encounters shows no difference in terms of service satisfaction~\cite{Soderlund20}. Until recent years, users have shown that human services are preferred over robot services, but due to the influence of COVID-19, robot services have also been shown to be particularly preferred in recent years~\cite{Kim21}. As in these studies, we can develop more valuable robots by not only showing that robots can accomplish tasks but also by comparing their performance with that of humans.

\section{Experiment I: With an Autonomous Robot}
\label{sec3}
This study aims to investigate whether the humanoid robot can make pedestrians stop and maintain engagement until they have delivered their intended message. To achieve this, three types of behavioral concepts were designed and compared.

For our investigations, we conducted an exploratory field experiment in a large shopping mall \footnote{Asia and Pacific Trade Center Co., Ltd. (ATC), Nankou-kita 2-1-10, Suminoe Area, Osaka-shi, Osaka, Japan} during July--August 2019. The humanoid robot was present for three weekdays and weekends and was available for 6 hours a day. The robot was placed in one of the shopping mall's corridors so that visitors, such as families, couples, and friends, could freely interact with the robot. We announced to all the pedestrians through a notification board that this was an experiment, and a video was being recorded along with the sensor data. This study was conducted on an opt-out basis for unwilling participants who wanted to be removed from the video and sensor data. The opt-out process may have changed the pedestrians' behavior, such as the pedestrians who attempted to interact with the robot, but quit the interaction owing to the notification. However, no one asked to delete the record in the experiment; thus, the effect of opt-out on the experimental results is expected to be minimal.

This experiment was approved by the facility authorities in the shopping mall and the Research Ethics Committee from Ritsumeikan University (Reference number: BKC-HitoI-2019-006).

\subsection{System Configuration}
\begin{figure}[!t]
  \centering
  \includegraphics[width=2.5in]{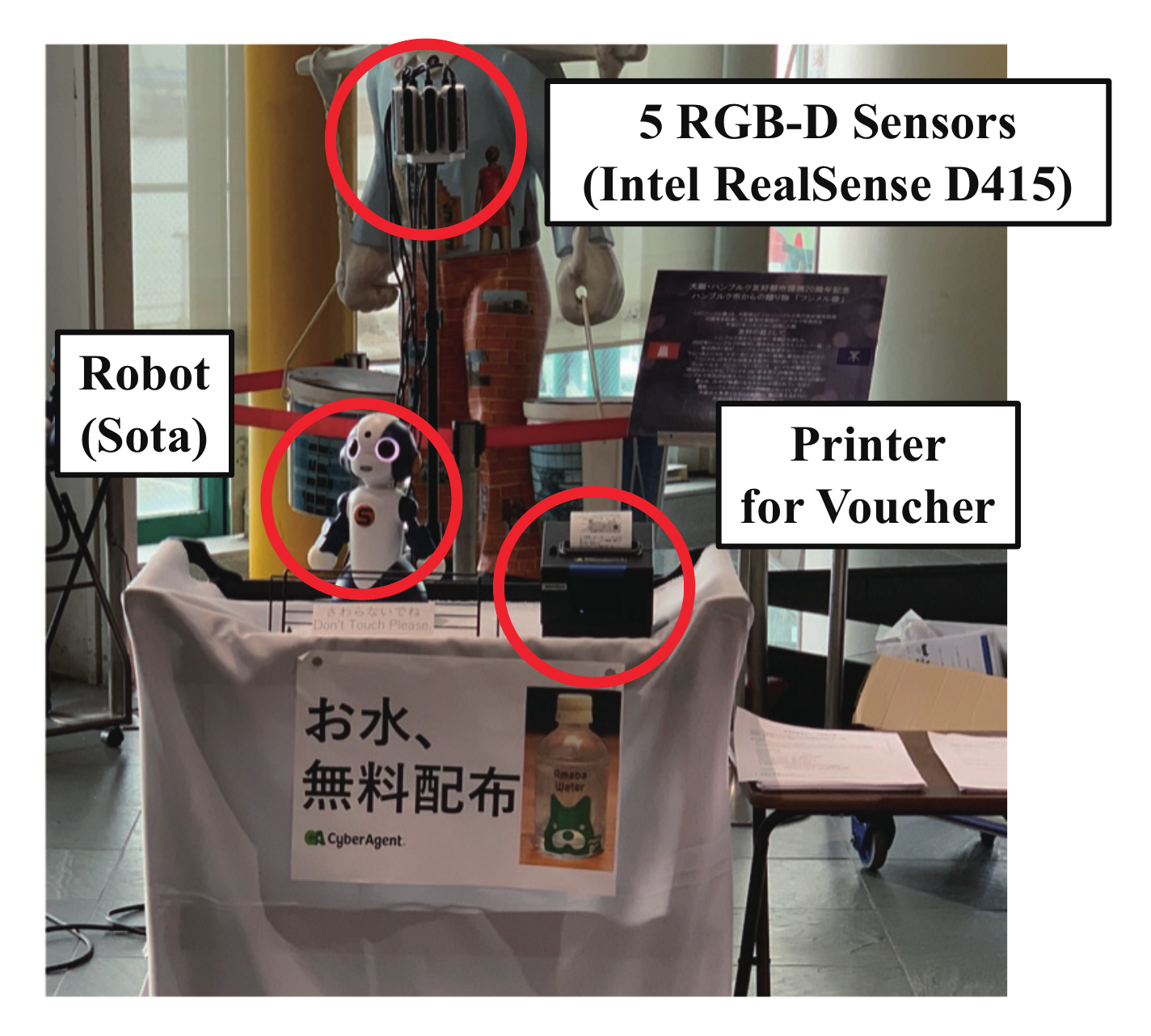}
  \caption{Interaction system. The robot and printer were installed on the desk, and the five RGB-D sensors were set behind the robot.}
  \label{fig:system}
\end{figure}
\begin{figure*}[!t]
  \centering
  \includegraphics[width=6.0in]{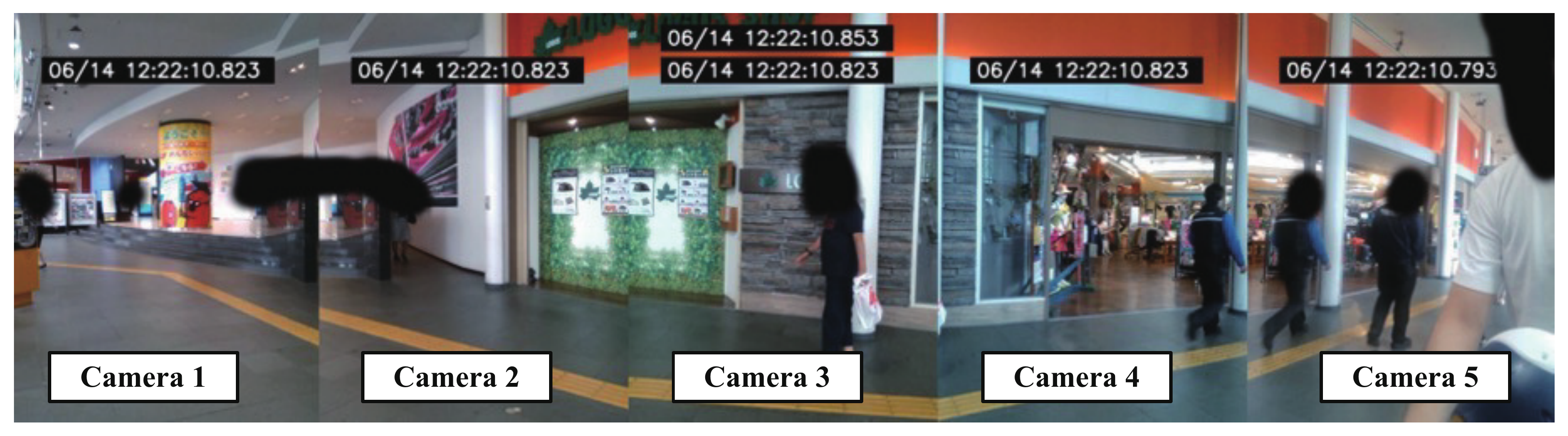}
  \caption{Combined image of the five RealSense cameras giving a total 220$^\circ$ FOV.}
  \label{fig:images}
\end{figure*}

We built an interaction system containing a humanoid social robot, five RGB-D image sensors, and a printer on a table, as shown in Fig.~\ref{fig:system}. The robot ``Sota'' that was developed by Vstone Co. Ltd. was used as the social robot in this experiment. The humanoid robot is approximately 0.3 m tall and has functions such as voice and LED-generated facial expressions. The robot includes arms with two degrees of freedom (DOF), head with three DOF, and body gestures with one DOF. Although the robot was equipped with a RGB camera on its head, we used additional cameras owing to the field of view (FOV) of the camera equipped on the robot, resulting in the limitations for these experiments. In terms of the 3D image sensors, we used five Intel RealSense D415 sensors, which can capture a RGB image (FOV per one camera: 69.4$^\circ$ $\times$ 42.5$^\circ$ $\times$ 77$^\circ$) and depth image (FOV per one camera: 65$^\circ$ $\times$ 40$^\circ$ $\times$ 72$^\circ$); the maximum range of the depth sensor was 10 m. A combined image of five cameras is shown in Fig.~\ref{fig:images}, and the horizontal FOV of the five sensors that were used in this study was 220$^\circ$. In addition, the printer was installed next to the robot to print a voucher for people who finished the interaction with the robot. The voucher could be exchanged for a bottle of water.

To generate robot behaviors according to the behavior of the pedestrians, we used the human detector ``NUITRACK,'' which can estimate the posture of the human in the image~\cite{NUITRACK}. From the results of NUITRACK and the depth image, we calculated the 3D coordinates of all the pedestrians that were observed from the robot's coordination. We used a computer (Intel Core i9-9900K CPU, NVIDIA GeForce RTX 2080 Ti GPU) to obtain the human posture of all pedestrians at a rate of 30 frames per second (fps). The limitation area to estimate the human posture was approximately within 4 m from the sensors.

\subsection{Interaction Design}
The robot had three types of behavioral concepts when pedestrians were near the robot: active, passive-negative, and passive-positive concepts. As mentioned in the Related Works section, this study aims to verify the behavioral concepts for the general use, rather than comparing the detailed differences for the behaviors. While exploring the behavioral concepts, we conducted a pre-experiment wherein we remotely controlled the robot at the university and performed a variety of behaviors to attempt to make pedestrians stop. From the results in the pre-experiment, we determined that there are three main methods that many people attempt during the task: active, passive-negative, and passive-positive.

During the active state, many operators that controlled the robot attempted to directly start a dialogue to the pedestrians such as ``Hello! How are you doing?'' or ``Where are you going to?''. Then, the conversation starts with the pedestrian replying to the robot. In contrast, some operators attempted other ways of dialogue initiated from the pedestrian rather than making the robot initiate a dialogue. For instance, the robot keeps muttering to itself ``I'm in trouble'' or dancing. The robot does not start the dialogue with the pedestrian until the pedestrian starts talking to the robot. It is important for the robot to generate an opportunity that makes pedestrians want to talk to the robot. Instead of starting a conversation when the pedestrian responds to the robot, the robot should make pedestrians feel that they have spoken to the robot. We call this method of starting the interaction from the pedestrian as a passive concept. In addition, we differentiated between the negative and positive expressions in the passive method. For instance, ``I'm in trouble'' and dancing were considered to be negative and positive expressions, respectively; thus, these are referred to as passive-negative and passive-positive concepts. In the pre-experiment, both passive behaviors demonstrated good results to attract pedestrians.

In summary, the behavioral concepts proposed in this study are as follows:
\begin{itemize}
  \item Active concept: A state in which a conversation starts when the pedestrian responds to the robot's talk.
  \item Passive-negative concept: A state in which the pedestrian observes the negative state of the robot and a conversation starts when the pedestrian first talks to the robot.
  \item Passive-positive concept: A state in which the pedestrian observes the positive state of the robot and a conversation starts when the pedestrian first talks to the robot.
\end{itemize}

This experiment here aimed to compare three behavioral concepts as the manipulative factors, rather than the differences in the detailed movements of the robots. We verified the influence of the behavioral concepts of the robot by measuring the behavior between participants at the same experimental location on different dates.

\subsection{Procedure}
According to the three proposed concepts that are based on the results of the pre-experiment, we specifically designed the ``greeting behavior'' as active, ``troubling behavior'' as passive-negative, and ``dancing behavior'' as passive-positive. The common robot motion for the three behaviors is face-to-face contact with the pedestrian that is closest to the robot so as to clarify whom the robot targets. Face-to-face contact has proven to be effective in human--robot interaction from various aspects, such as conveying the robot's attention~\cite{Admoni17}. For the greeting behavior, the robot makes hand raising gestures to the pedestrians and says ``Hello! Please talk with me!'' During the troubling state, the robot behaves as if it has a headache and keeps muttering to itself for seeking someone's help ``I'm in trouble. What should I do?'' This creates an opportunity for the pedestrian to want to talk to the robot by pretending that the robot is in trouble. During the dancing mode, the robot dances while singing. This also creates an opportunity for the pedestrian to want to talk to the robot for fun.

These behaviors were performed while the posture for at least one pedestrian was measured (maximum range to obtain the postures was 4 $\times$ 4 m as the lateral and depth direction) until the pedestrian stopped in front of the robot. The robot determined if the pedestrian stopped by examining whether or not the pedestrian stayed in the area of 1.2 $\times$ 2.5 m (lateral and depth direction) from the sensors for 3 s. After the pedestrian stopped in front of the robot, the robot started talking about a store in the shopping mall for 13, 19, or 26 s, depending on the scenario. This talk scenario was generated randomly regardless of the three types of behaviors. While the robot was talking, the interaction system with the passersby was not performed, which is called a passive medium~\cite{Hayashi07}. When the pedestrian finished listening to the robot, the printer next to the robot printed a voucher. The pedestrian could exchange the voucher for a bottle of water in the store that the robot was advertising.

These series of robot behaviors were all performed automatically. As explained above, the robot behavior is not determined by verbal interaction with the pedestrian. This is because the environment of the commercial facility was noisy, and the robot system could not accurately recognize what the pedestrians talk to the robot. Therefore, the robot behavior was automatically generated by estimating the pedestrian's state based only on the posture data. Each behavior in motion can be watched from a video that was presented in the preliminary study~\cite{youtube}.

\subsection{Hypothesis}
The greeting behavior was considered a basis because this is often used to get the attention of pedestrians (e.g.,~\cite{Saad19}). On the other hand, previous studies have demonstrated that emotional robots~\cite{Leite08} and human-dependent robots~\cite{Khaoula14}, which are similar to the idea of troubling (passive-negative), can have a higher engagement with users. Therefore, we can expect that troubling also demonstrates a higher performance in this experiment. In addition, the high performance by the dancing behavior (passive-active) is also expected because dancing for rhythmic interaction was performed with higher engagement with children~\cite{Michalowski07}.
\begin{figure}[!t]
  \centering
  \includegraphics[width=3in]{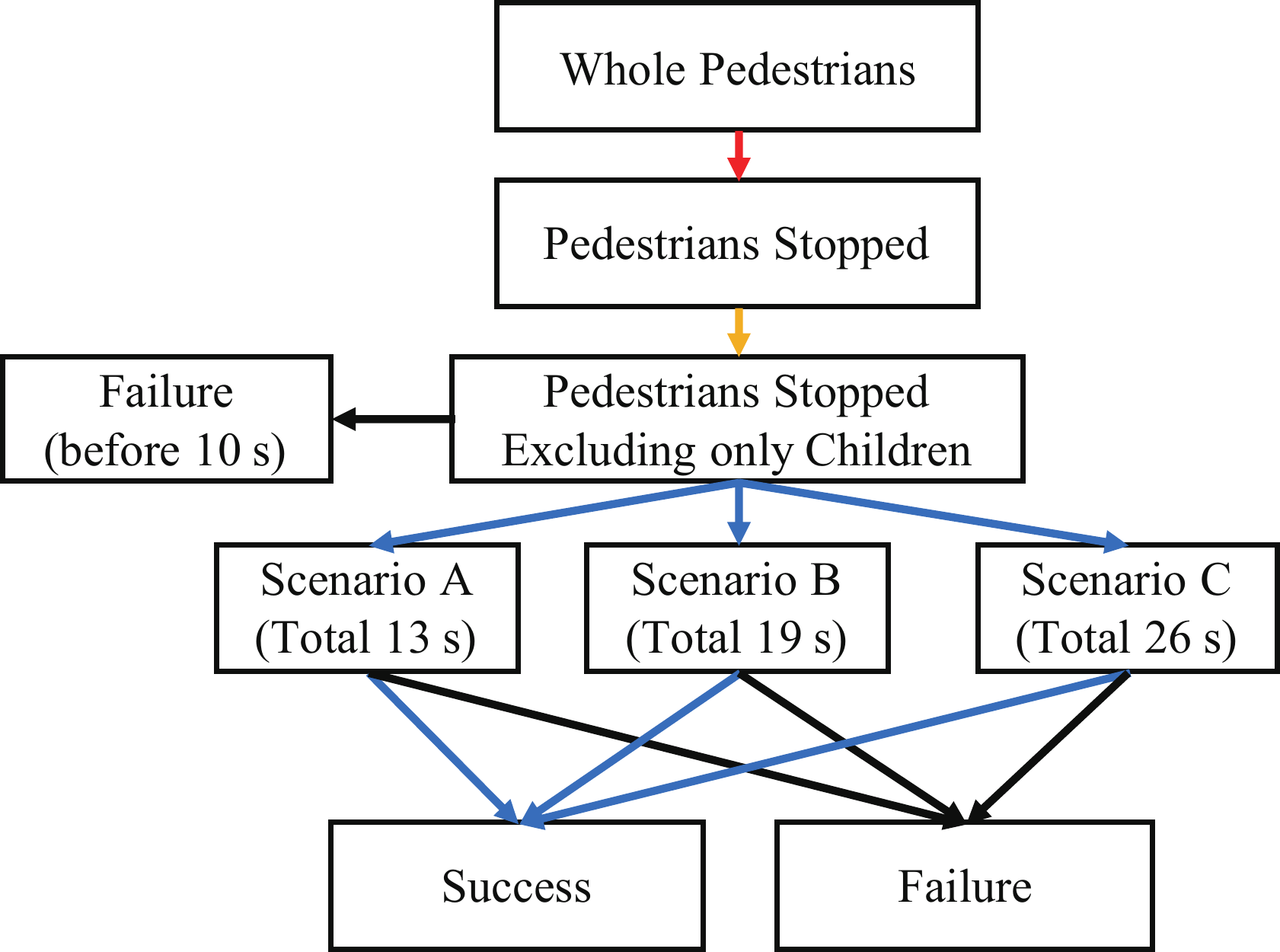}
  \caption{Paths of all the labels that were annotated from the recorded video. The red and blue paths represent the stop rate (SR) and the distribution success rate (DSR). The path through red-yellow-blue denotes the whole distribution success rate (WDSR).}
  \label{fig:label}
\end{figure}
\begin{table*}[!t]
  \renewcommand{\arraystretch}{1.3}
  \caption{Total number of pedestrians that walked in front of the robot, pedestrians who stopped in front of the robot, pedestrians stopped while excluding only children, and pedestrians who received the voucher}
  \label{tab:results1}
  \centering
  \begin{tabular}{llcccc}
    \hline
    \multicolumn{2}{c}{Conditions} & Whole Pedestrians & Pedestrians Stopped & Pedestrians Stopped excluding only children & Pedestrians Received \\ \hline 
    \multirow{2}{*}{Greeting}  & Weekday & 6306  & 362  & 273 & 94  \\
                               & Weekend & 19738 & 1139 & 925 & 312 \\
    \multirow{2}{*}{Troubling} & Weekday & 6000  & 559  & 443 & 186 \\
                               & Weekend & 10205 & 1162 & 953 & 396 \\
    \multirow{2}{*}{Dancing}   & Weekday & 7509  & 819  & 644 & 169 \\
                               & Weekend & 16884 & 1057 & 819 & 211 \\
    \hline
  \end{tabular}
\end{table*}

\subsection{Measurement}
Throughout the experiment, we recorded videos as shown in Fig.~\ref{fig:images} to analyze the behavior of all the pedestrians. Thereafter, we labeled all the pedestrians that walked in front of the robot front, the pedestrians that stopped in front of the robot, and the pedestrians who received the voucher. Along with counting the people that stopped in front of the robot, we labeled pedestrians that stopped while excluding situations where there were only children. The reason for excluding only children from the ``pedestrian stopped'' situation was that the human detector NUITRACK cannot detect small children due to the occlusion by the robot and the desk. Thus, the proposed robot system was not able to proceed to the information provision phase even if the children were in front of the robot. The number of people received was labeled depending on the talk scenarios (A: 13 s, B: 19 s, and C: 26 s). The detailed path of all labels is shown in Fig.~\ref{fig:label}.

The features extracted from the recorded videos include the labeled behavior, apparent gender of the pedestrian(s), and estimated age (child under 12 years old or an adult). This annotation was applied to all the pedestrians who passed by the robot in the video. If the same person appeared more than once a day, each appearance was annotated without personal identification. To ensure valid results, video data annotation was performed by two coders. One was the author, Y. O., and the other was a person unrelated to this study, who was hired as a part-time worker. In order to ensure uniformity in the criteria while judging ``stop'' behaviors in front of the robot, the two coders worked together to determine the criteria before annotating. The data for one day were overlapped, and the analysis of the overlapped data showed that they were well matched (Cohen's Kappa was .894).

\begin{figure*}[!t]
  \centering
  \subfloat[SR]{\includegraphics[width=2.3in]{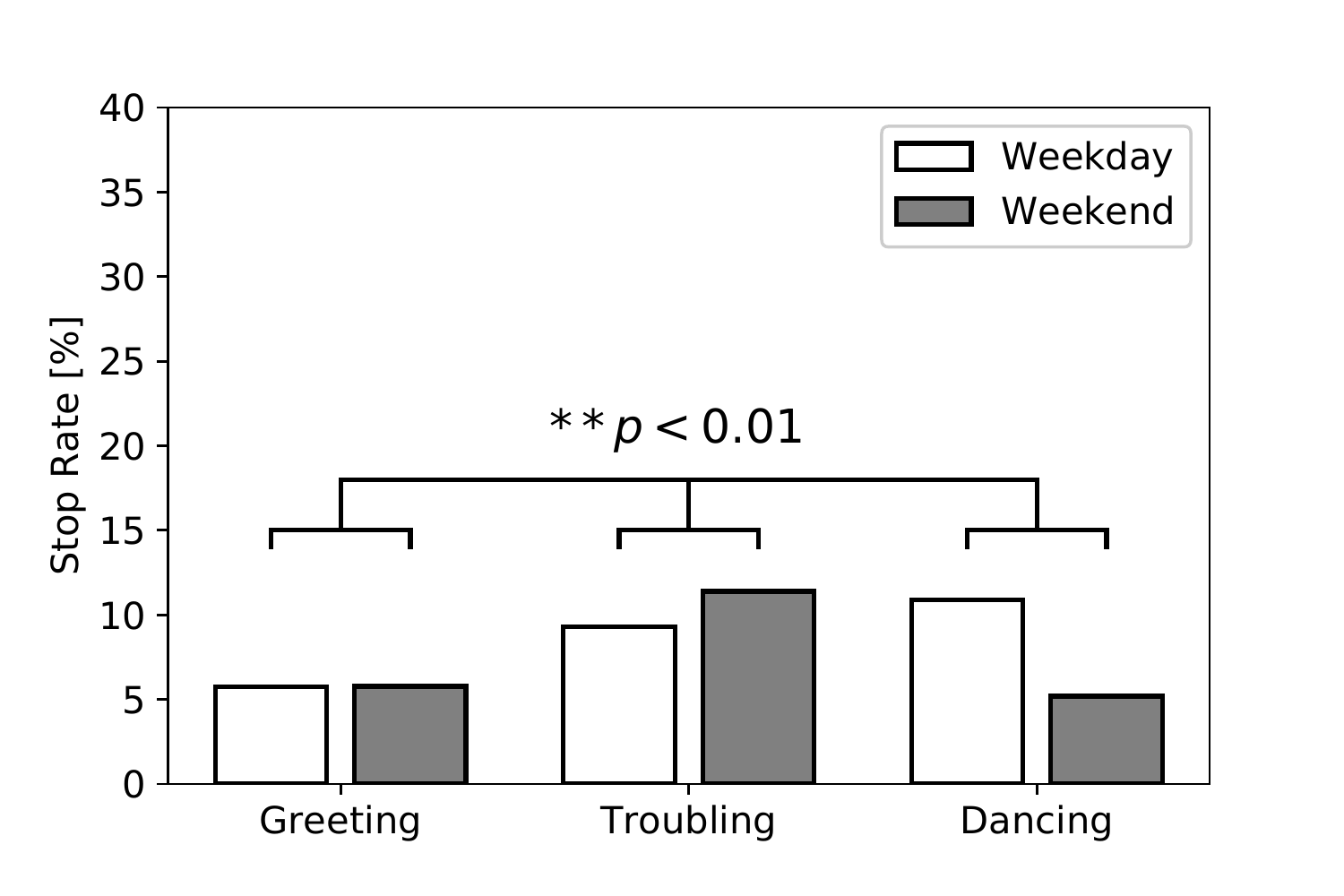}
    \label{fig:SR_robot}}
  \hfil
  \subfloat[DSR]{\includegraphics[width=2.3in]{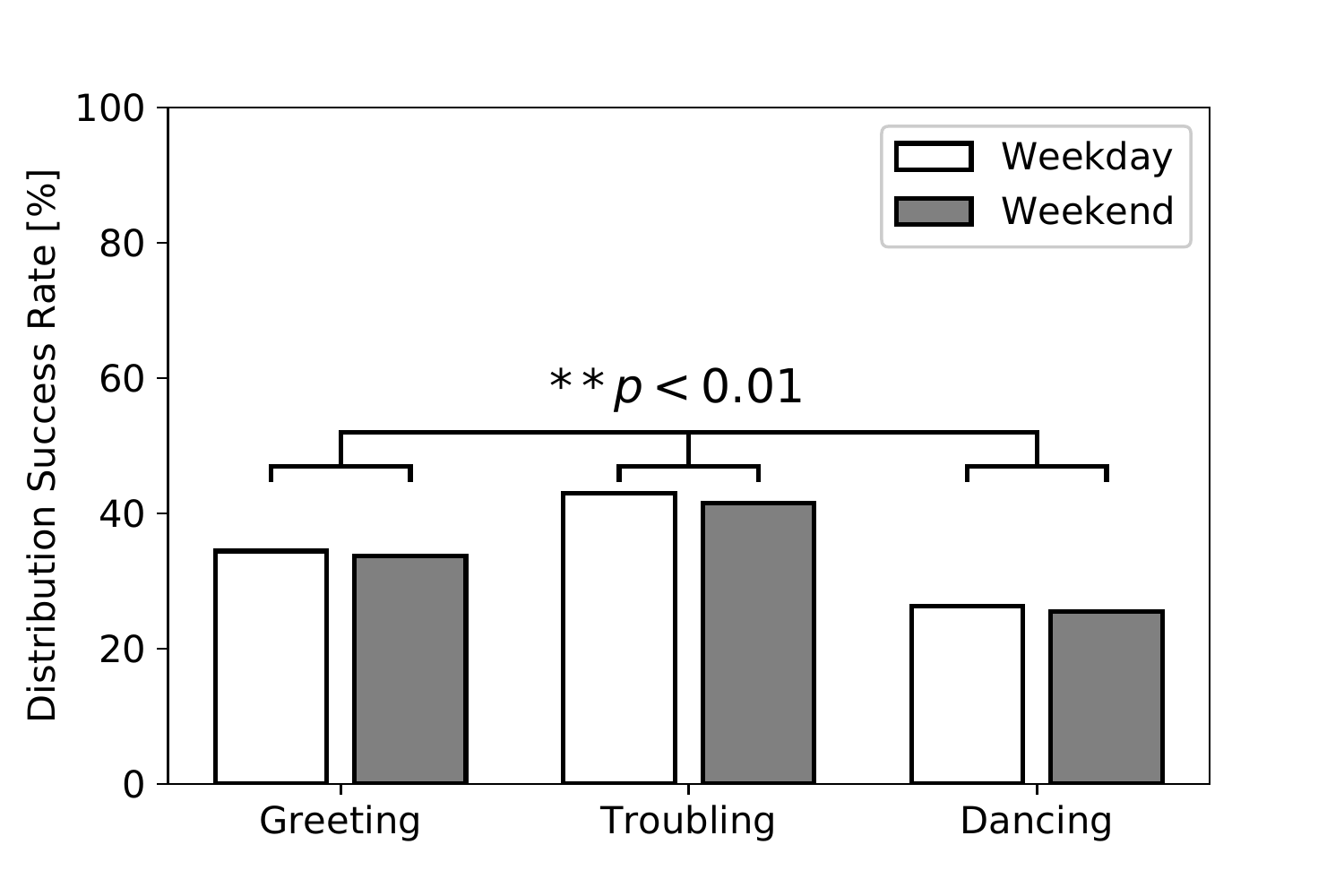}
    \label{fig:DSR_robot}}
  \hfil
  \subfloat[WDSR]{\includegraphics[width=2.3in]{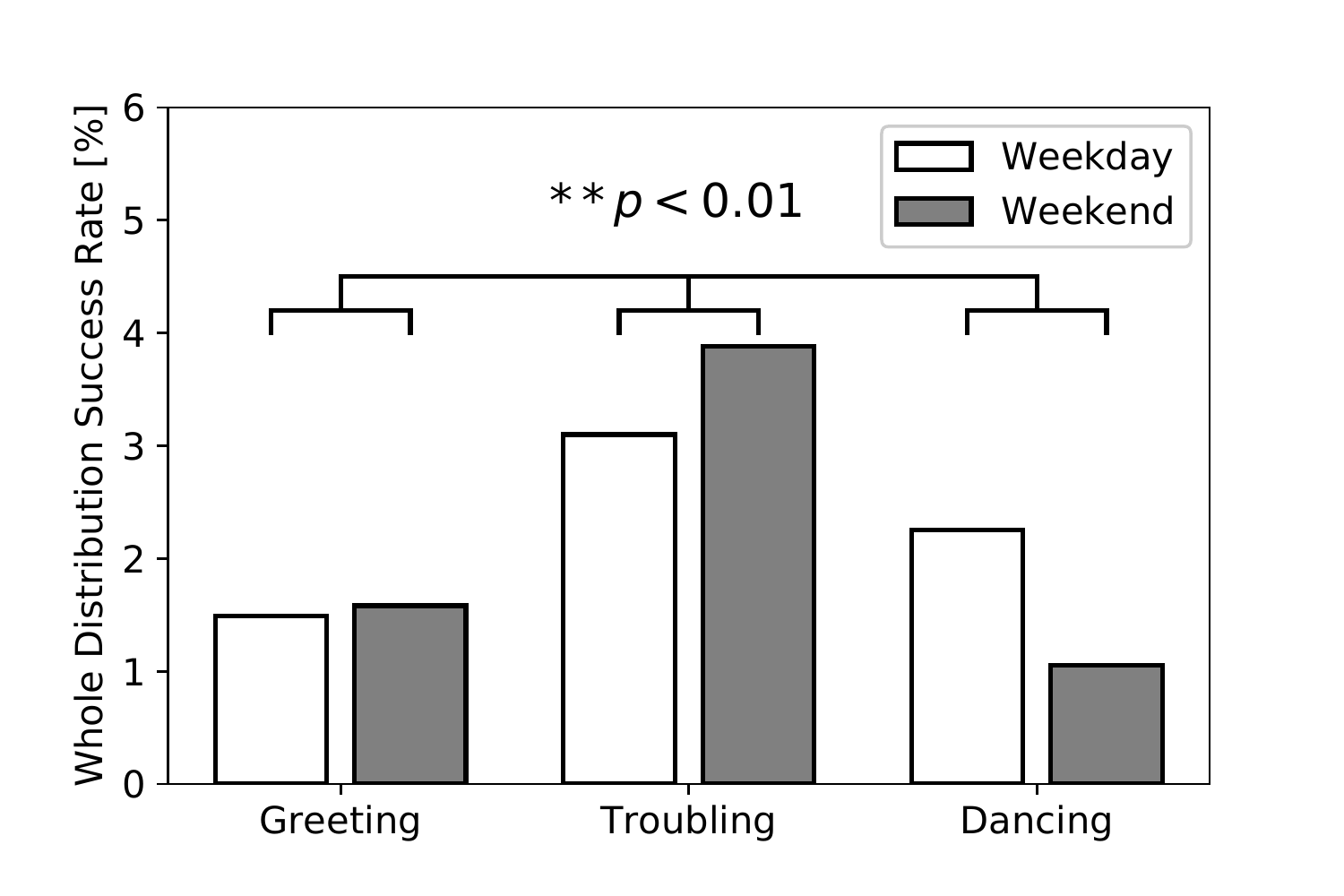}
    \label{fig:WDSR_robot}}
  \caption{Results of the stop rate (SR), distribution success rate (DSR), and the whole distribution success rate (WDSR) according to each robot behavior in Experiment I.}
  \label{fig:whole_robot}
\end{figure*}

We used three indexes for the evaluation: the stop rate (SR), distribution success rate (DSR), and whole distribution success rate (WDSR). The SR is the ratio of the number of pedestrians that stopped to the number of all pedestrians (red path in Fig.~\ref{fig:label}). The DSR is the ratio of the number of pedestrians who received the voucher to the pedestrians that stopped, while excluding only children (blue path in Fig.~\ref{fig:label}). The WDSR is the ratio of the number of pedestrians who received the voucher to the number of all pedestrians (path through red-yellow-blue in Fig.~\ref{fig:label}).

\subsection{Results}
\begin{figure*}[!t]
  \centering
  \subfloat[SR]{\includegraphics[width=2.3in]{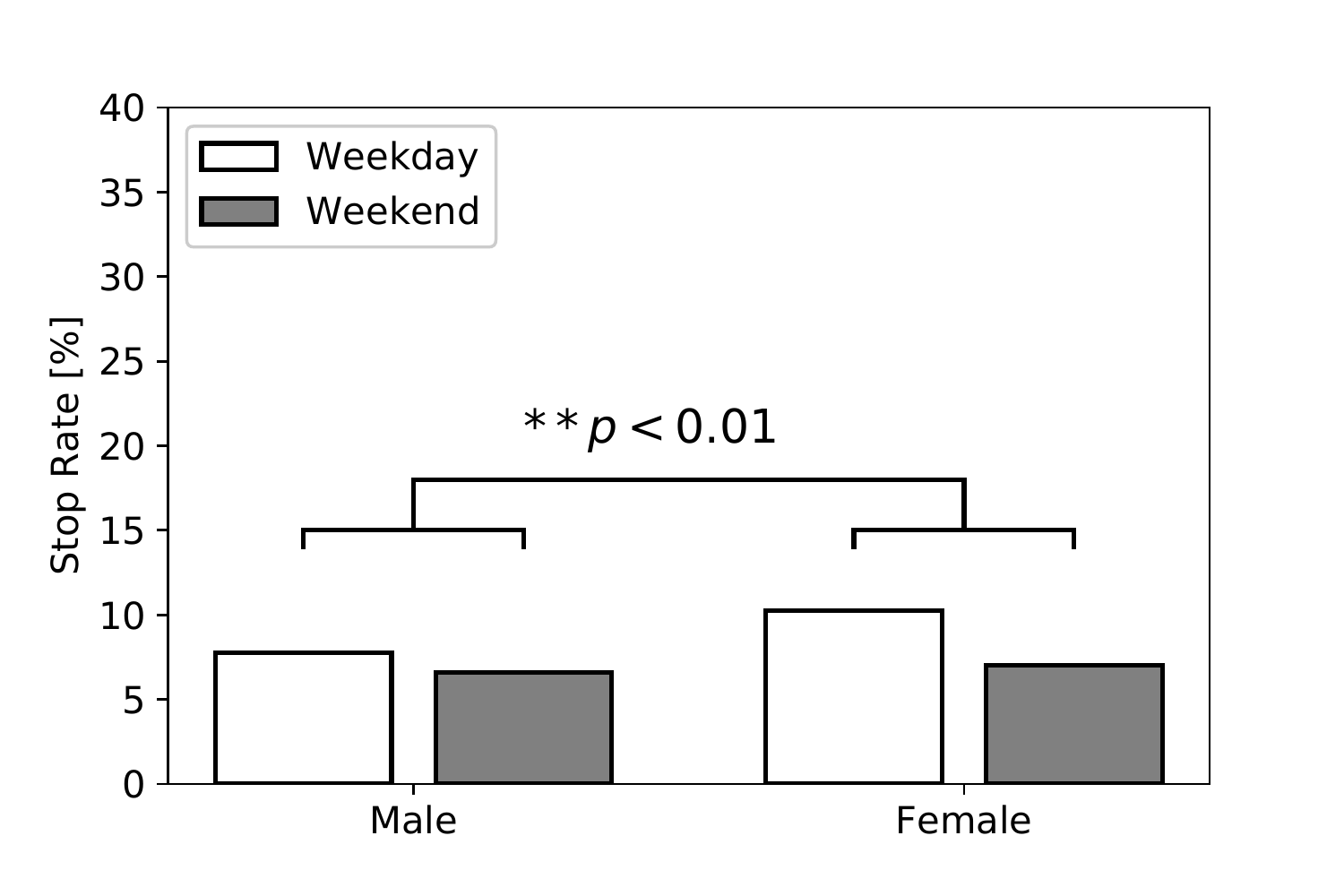}
    \label{fig:gender_SR_robot}}
  \hfil
  \subfloat[DSR]{\includegraphics[width=2.3in]{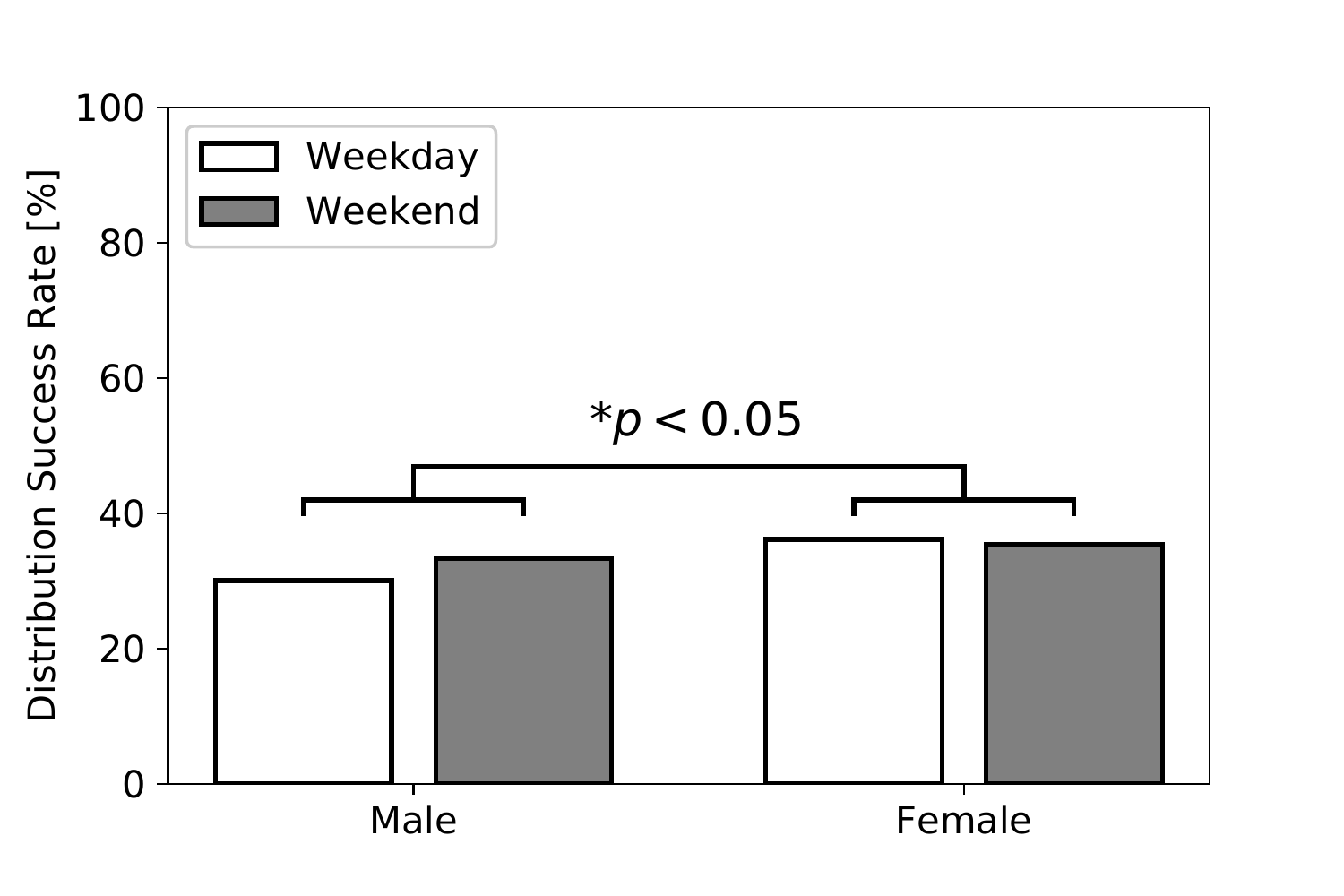}
    \label{fig:gender_DSR_robot}}
  \caption{Results of the stop rate (SR) and the distribution success rate (DSR) according to the gender in Experiment I.}
  \label{fig:gender_robot}
\end{figure*}

\begin{figure*}[!t]
  \centering
  \subfloat[SR]{\includegraphics[width=2.3in]{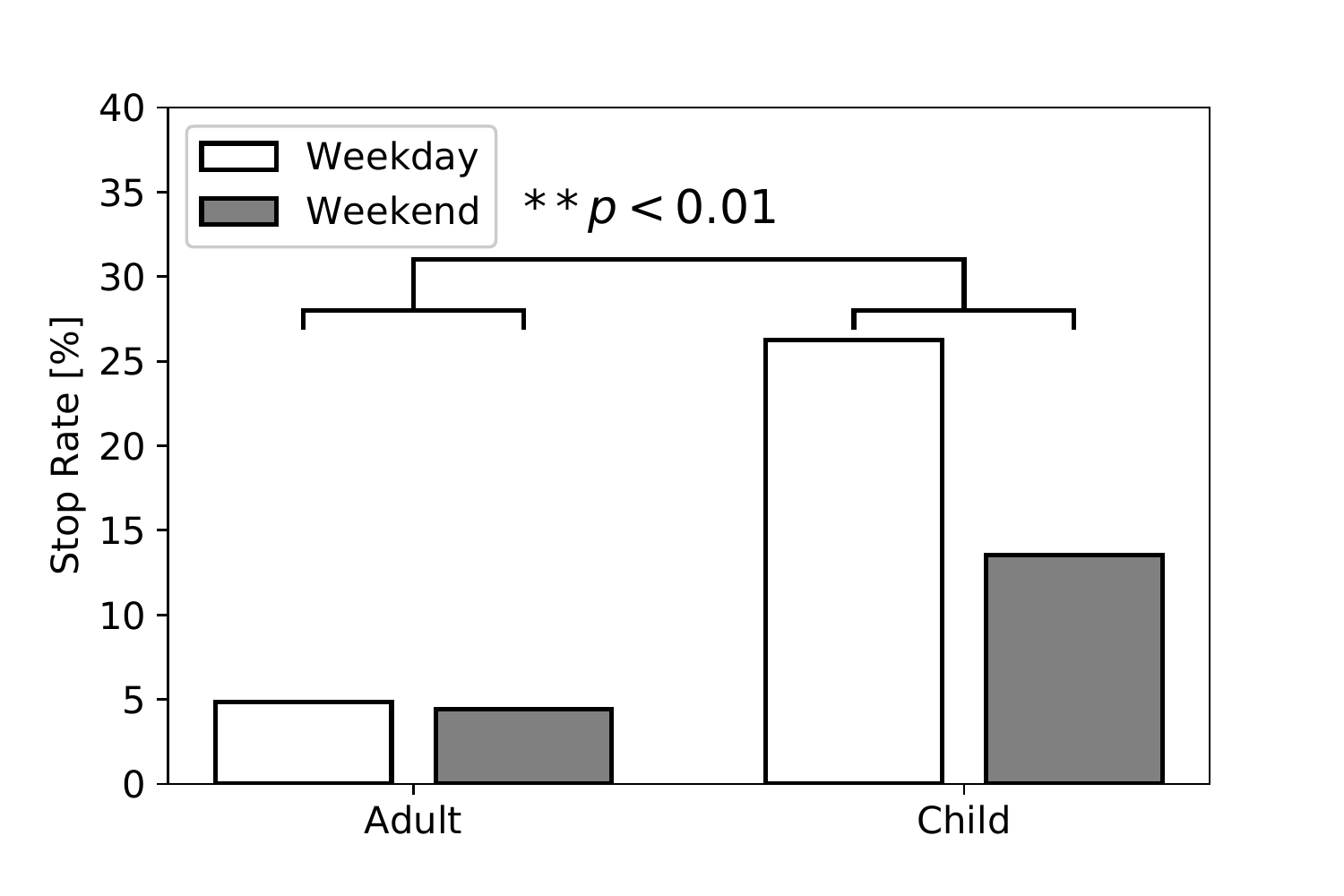}
    \label{fig:age_SR_robot}}
  \hfil
  \subfloat[DSR]{\includegraphics[width=2.3in]{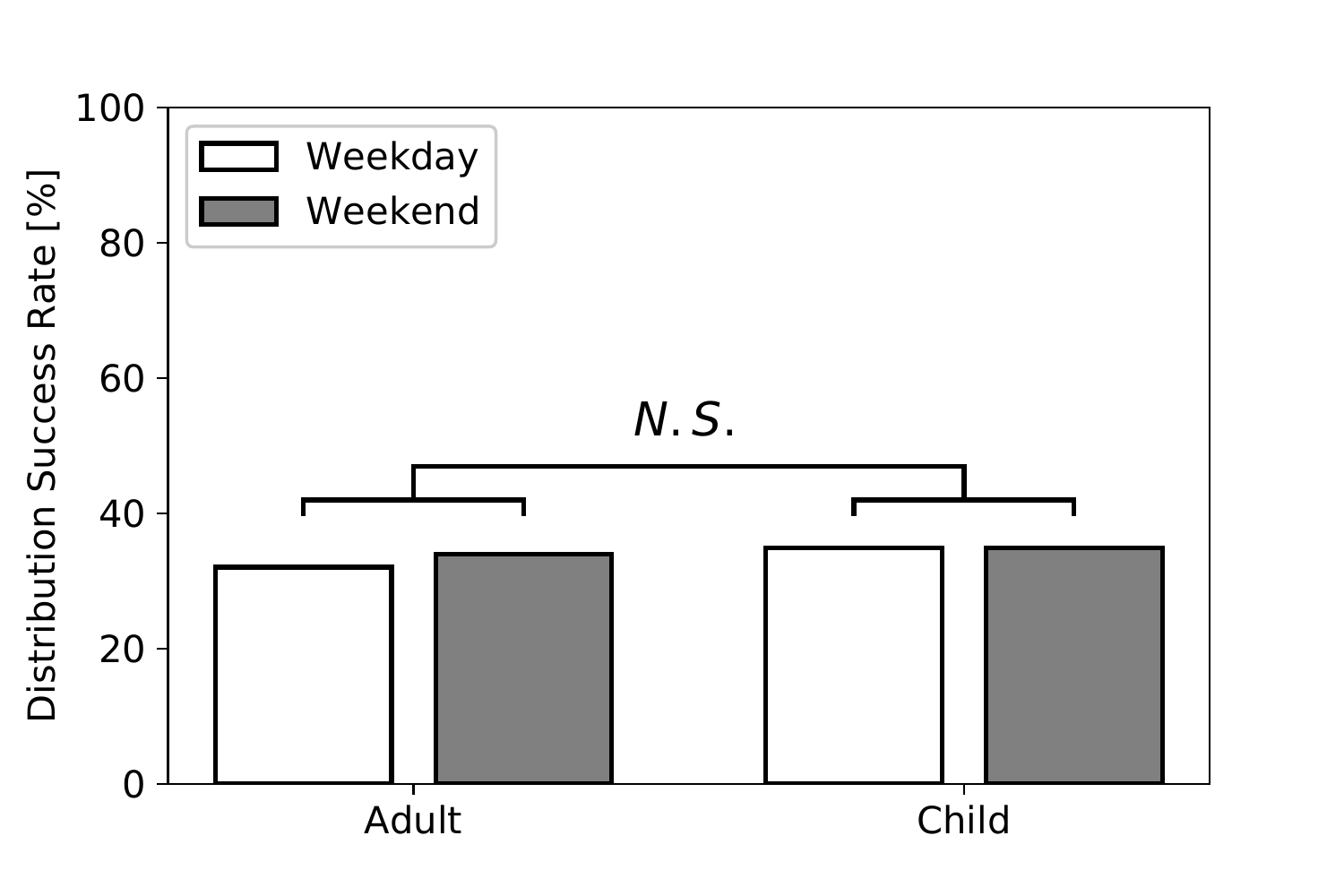}
    \label{fig:age_DSR_robot}}
  \caption{Results of the stop rate (SR) and the distribution success rate (DSR) according to the age in Experiment I.}
  \label{fig:age_robot}
\end{figure*}

\begin{figure}[!t]
  \centering
  \includegraphics[width=2.5in]{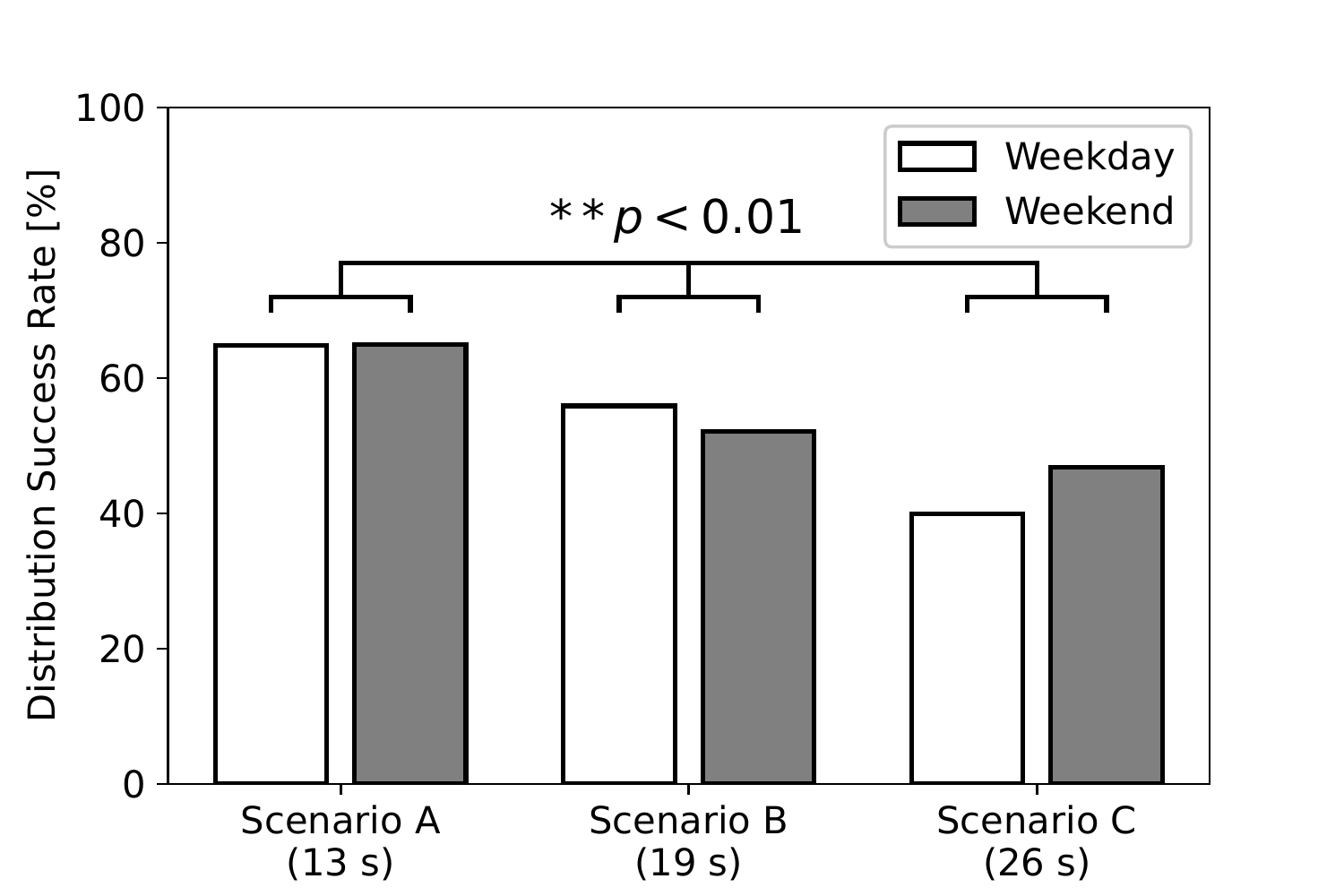}
  \caption{Results of the distribution success rate (DSR) according to each scenario in Experiment I.}
  \label{fig:scenario_robot}
\end{figure}

\begin{figure*}[!t]
  \centering
  \subfloat[SR]{\includegraphics[width=2.4in]{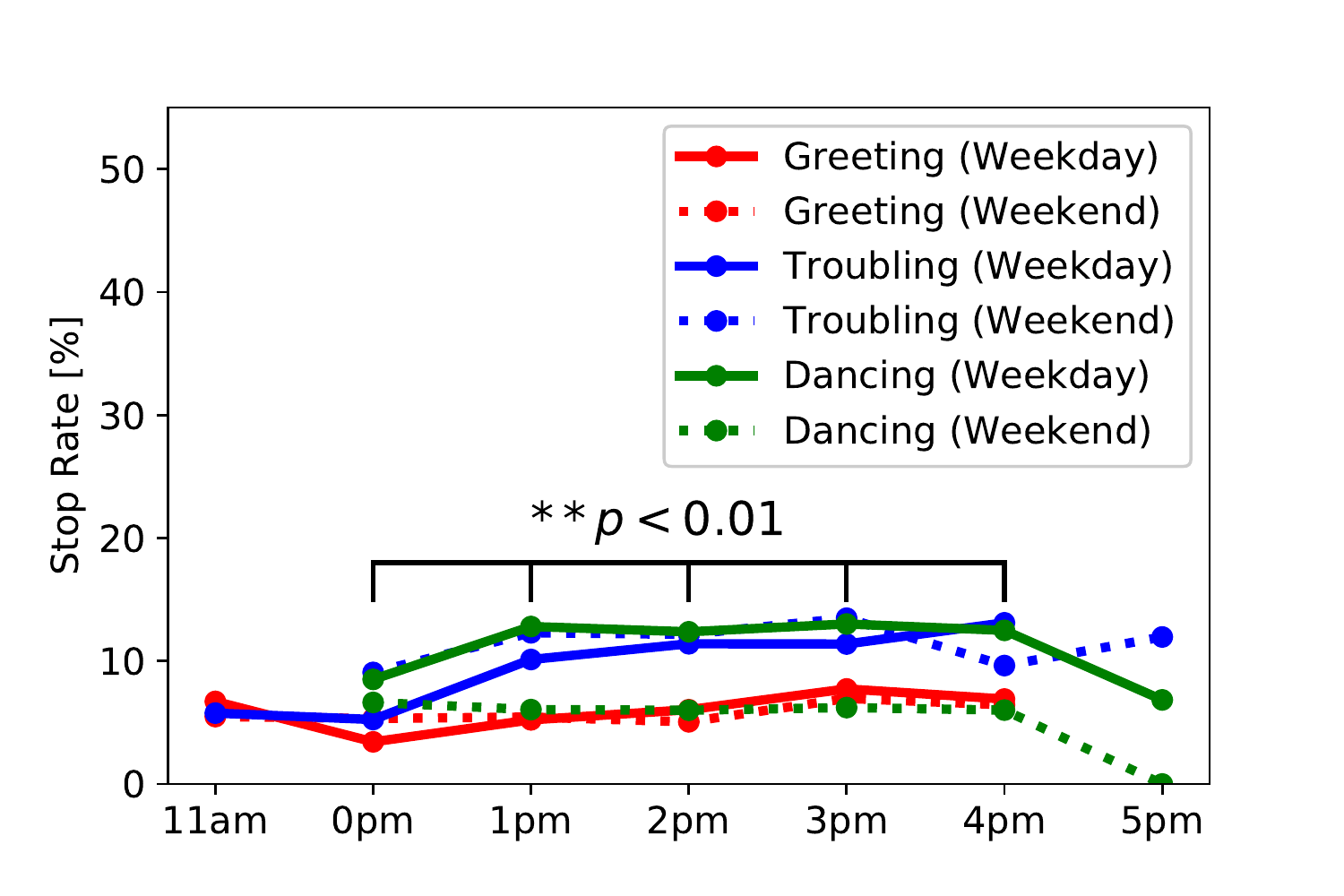}
    \label{fig:time_SR_robot}}
  \hfil
  \subfloat[DSR]{\includegraphics[width=2.4in]{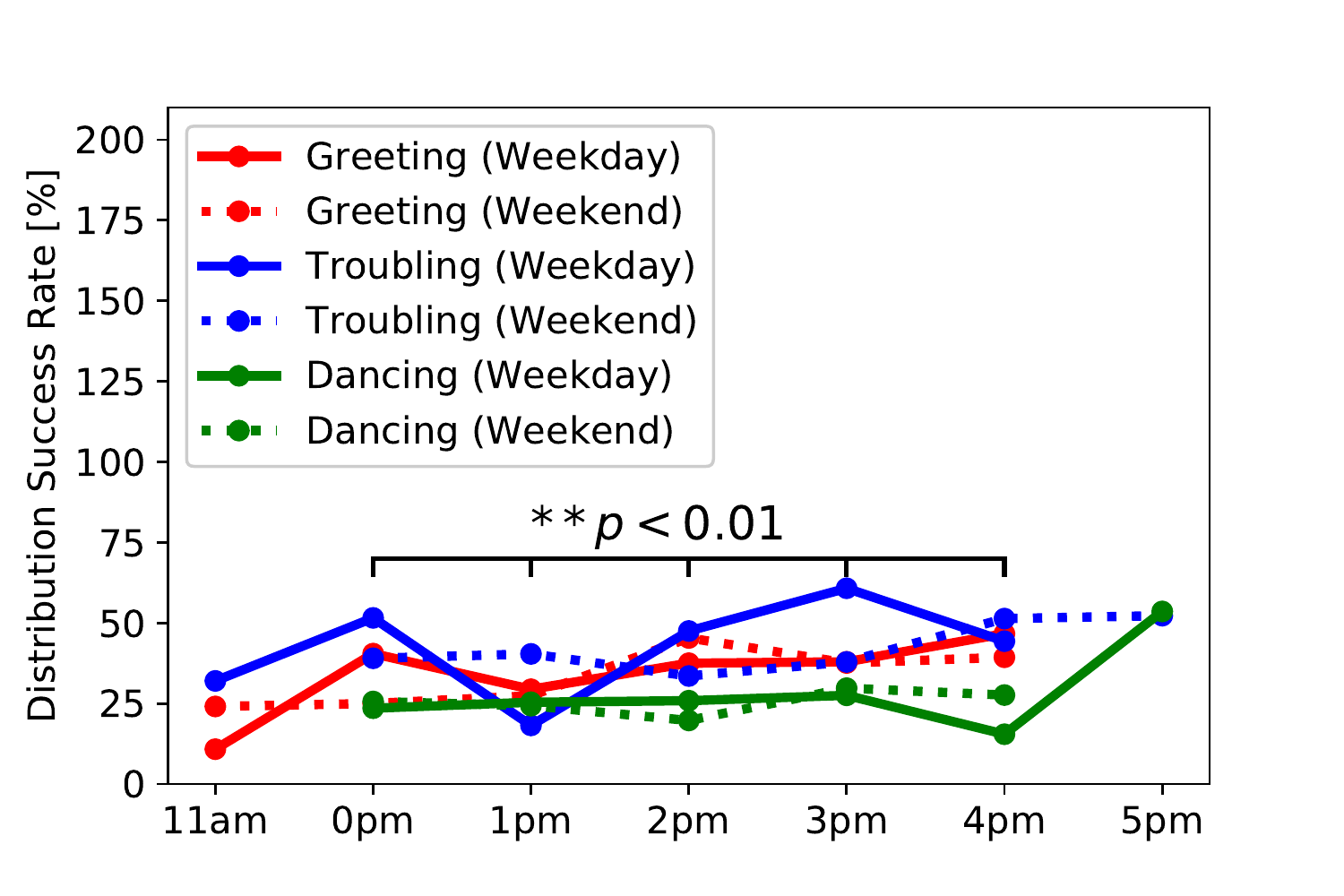}
    \label{fig:time_DSR_robot}}
  \caption{Results of the stop rate (SR) and the distribution success rate (DSR) according to the time window in Experiment I.}
  \label{fig:time_robot}
\end{figure*}

\begin{table*}[!t]
  \renewcommand{\arraystretch}{1.3}
  \caption{Ratio of the number of times between one pedestrian and the group that stopped in front of the robot and listened to the robot's message until the end.}
  \label{tab:results2}
  \centering
  \begin{tabular}{llcccc}
    \hline
    \multicolumn{2}{c}{Conditions} & \multicolumn{2}{c}{Ratio of pedestrians stopped} & \multicolumn{2}{c}{Ratio of pedestrians received} \\
    & & One pedestrian \% & Two or more pedestrians \% & One pedestrian \% & Two or more pedestrians \% \\ \hline
    \multirow{2}{*}{Greeting}  & Weekday & 36.2 & 63.8 & 13.5 & 86.5 \\
                               & Weekend & 35.1 & 64.9 & 11.0 & 89.0 \\
    \multirow{2}{*}{Troubling} & Weekday & 39.5 & 60.5 & 10.0 & 90.0 \\
                               & Weekend & 34.3 & 65.7 & 8.3  & 91.7 \\
    \multirow{2}{*}{Dancing}   & Weekday & 37.3 & 62.7 & 12.5 & 87.5 \\
                               & Weekend & 38.1 & 61.9 & 10.0 & 90.0 \\
    \hline
  \end{tabular}
\end{table*}

The results for the number of labeled pedestrians are shown in Table~\ref{tab:results1}. The results for the SR, DSR, and WDSR according to each robot behavior are shown in Fig.~\ref{fig:whole_robot}. We verified the differences in the number of pedestrians among the behavior conditions with a Chi-squared test. We used the Cramer's $V$ as the effect size in all the Chi-square test results. The results revealed significant differences among the behavioral conditions: $(\chi^2(2) = 333.64, p < 0.01, V = 0.05)$ in the SR, $(\chi^2(2) = 81.61, p < 0.01, V = 0.10)$ in the DSR, and $(\chi^2(2) = 252.14, p < 0.01, V = 0.04)$ in the WDSR. The residual analysis in comparison with the mean across all behaviors showed that (1) the greeting had low SR and WDSR ratios, (2) the troubling state had high SR, DSR, and WDSR ratios, and (3) dancing had a high SR ratio but low DSR and WDSR ratios. Therefore, our results show that the robot motion that behaves as though it is in trouble makes pedestrians stop more and stay longer in front of the robot in comparison to the greeting and dancing behaviors. Meanwhile, the dancing behavior resulted in a significant number of pedestrians to stop, but the stopped pedestrians did not listen to the robot talk for a long time. 

As described in the detailed analysis, the additional results of the gender difference in the SR and DSR, age difference in the SR and DSR, and scenario difference in the DSR are shown in Figs.~\ref{fig:gender_robot}-\ref{fig:scenario_robot}. We also verified the differences in the number of people for the gender in the SR and DSR, age in the SR and DSR, and scenario in DSR through a Chi-square test. The results among the gender revealed significant differences in the SR $(\chi^2(1) = 21.50, p < 0.01, V = 0.02)$ and the DSR $(\chi^2(1) = 5.26, p = 0.02, V = 0.04)$. The results among the age showed significant differences in the SR $(\chi^2(1) = 2666.67, p < 0.01, V = 0.20)$, but no significant differences in the DSR $(\chi^2(1) = 1.10, p = 0.29, V = 0.02)$. As a result, this indicates that females and children are more likely to stop in front of the robot. However, there were no gender or age differences in whether they listened to the message to the end. In terms of the talk scenario difference (Scenario A: 13 s, Scenario B: 19 s, Scenario C: 26 s), there was a significant difference in the DSR $(\chi^2(2) = 77.37, p < 0.01, V = 0.12)$. In addition, the findings revealed that the pedestrians listen to the robot talk in its entirety as the scenario becomes shorter.

Next, there were differences in pedestrian behavior in the hourly results; Fig.~\ref{fig:time_robot} shows the SR and DSR results for each hour. As the start time of the experiment was different for each condition, we validated the results from 12--4 pm, which is the common time window when the experiment was conducted, with a Chi-square test. The results among the time revealed significant differences in the SR $(\chi^2(4) = 65.76, p < 0.01, V = 0.02)$ and the DSR $(\chi^2(4) = 21.28, p < 0.01 , V = 0.04)$. The residual analysis in comparison with the mean across all times showed that (1) 12 pm had low SR ratio, (2) 1 pm had low DSR ratio, and (3) 3 pm and 4 pm had high SR and DSR ratios. Therefore, these results indicate that it is difficult for robots to approach pedestrians during the hours close to lunchtime.

Finally, we show the ratio of the number of times between one pedestrian and the group that stopped in front of the robot and listened to its message until the end, as shown in Table~\ref{tab:results2}. The results revealed that the differences are small for the behavioral conditions in terms of the pedestrians stopping rates and the received pedestrians. In addition, it was determined that pedestrians in a group stop in front of the robot and listen to its entire message in comparison to those who are alone.

\section{Experiment II: By Humans}
\label{sec4}
Experiment I aims to identify robot behaviors that can make pedestrians stop and maintain engagement with them. The results show that the troubling behavior can attract more pedestrians based on a high SR, DSR, and WDSR. Meanwhile, the results of the troubling behavior are based on the relative results in comparison to the two other types of robot behaviors. Therefore, we cannot argue that these robot behaviors are more competent than human behaviors. If robots are to be used as a labor support technology, a better performance by robots then humans is desirable. In addition, clarifying the tasks wherein robots are superior to humans provides invaluable insights in collaborating with humans. Several studies have been conducted to examine whether robots can play an active role in the real world by comparing the performance of robots and humans (e.g.~\cite{Watanabe15}). Thus, to compare the results that are generated by the robots and those from humans, we conducted a second experiment in which the robot was replaced by a person in an equivalent experimental environment.

The experiment was conducted in the same location as Experiment I in November 2019. With the same situation as Experiment I, the experiment was recorded with a notification board and conducted on an opt-out basis.

This experiment was approved by the facility authorities in the shopping mall and the Research Ethics Committee from Ritsumeikan University (Reference number: BKC-HitoI-2019-006-1).

\subsection{Human Advertisers}
Four people who have experience in distributing flyers were recruited through a temporary agency to participate in the experiment (two males/females, average age: 23.75 years old). The experiment was conducted over 4 days, and each human advertiser performed the task each day. All the human advertisers provided informed consent, which allowed for the use of the collected data for scientific purposes and publication. The human advertisers received 9,500 JPY a day, and they also received a reward in accordance to their performance.

\subsection{Interaction Design}
The human advertisers were instructed to encourage the pedestrians to stop to provide information about the shop. The advertisers were allowed to use the leaflet of the shop, which is different from Experiment I. This task requires the human advertisers to perform their usual way of providing information. As one of the goals in Experiment II was to compare the robot's performance with the human's usual performance, human behavior should be as unrestricted as possible. After the pedestrians were stopped by the advertiser's attempt, the advertiser must provide them with information about the shop, such as recommended products. It was forbidden to annoy the pedestrians to force them to stop, such as interfering with their walking or saying that water is being distributed. When the pedestrian finished listening to the information about the shop, the advertiser gave a voucher that could be exchanged for a bottle of water. 

As a standard for the area in which the advertiser could move during the experiment, an area of 1.2 $\times$ 2.5 m (lateral and depth direction) was specified, which is the same area for determining the pedestrians to stop in Experiment I. Thus, the significant difference between Experiments I and II was whether to use the leaflet, and other situations were set up to be similar. However, we treated Experiments I and II as separate experiments because the conditions were not completely consistent.

\begin{figure}[!t]
  \centering
  \includegraphics[width=2.5in]{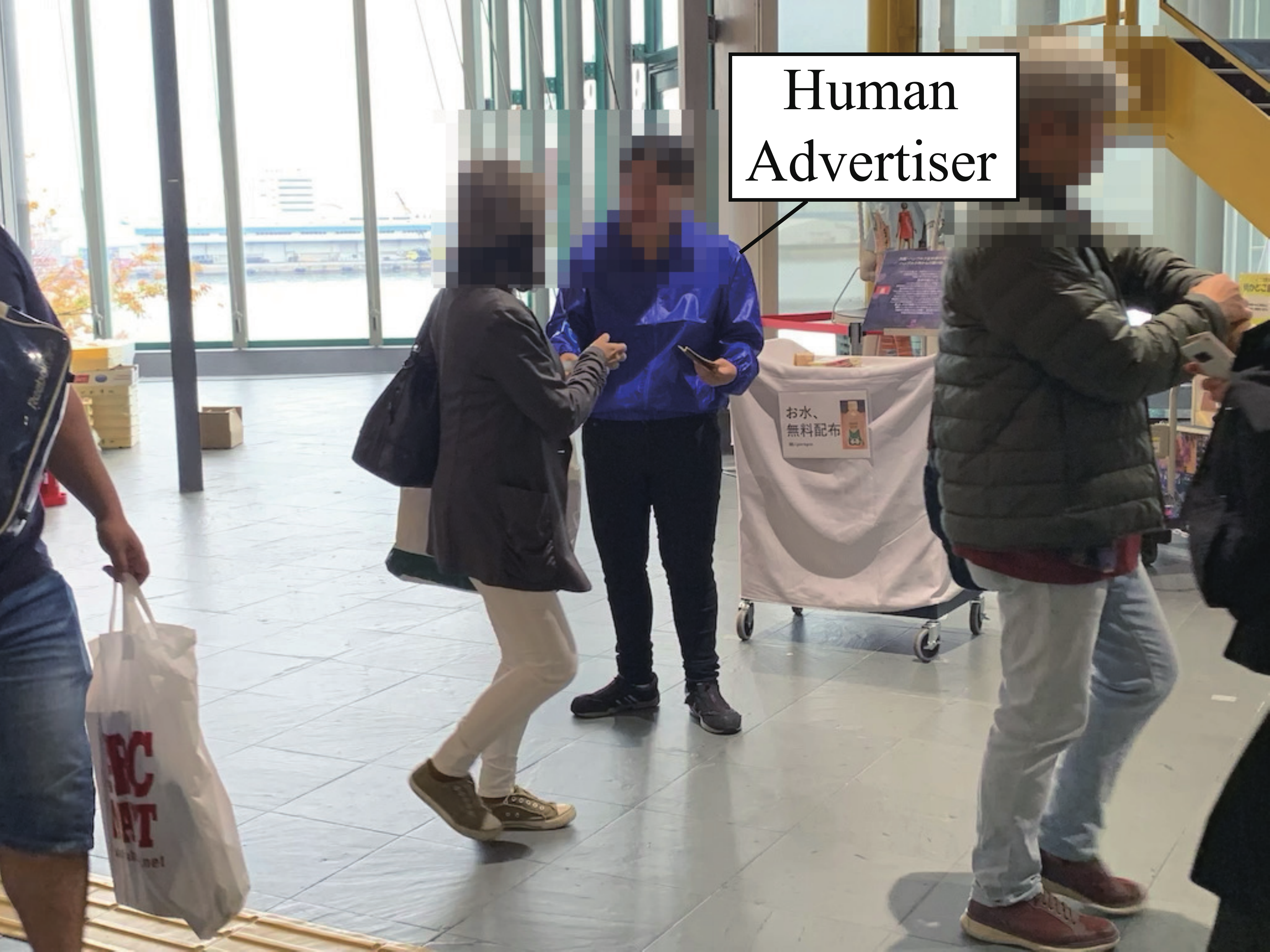}
  \caption{Example scene in Experiment II with a human advertiser conveying information to pedestrians.}
  \label{fig:Ex2}
\end{figure}

The human advertisers were allowed to practice the experimental task for approximately 20 min before the experiment. The advertisers performed three sets of 50 min of executing the experimental task followed by 10 min of rest. To provide motivation to the advertisers, an additional reward of 500 JPY for every 10 vouchers were given to the advertisers. The scene during Experiment II is shown in Fig.~\ref{fig:Ex2}.

\subsection{Measurement}
We labeled some indices from the recorded video, which are similar to the labels in Experiment I, as shown in Fig.~\ref{fig:label}. While the robot system in Experiment I could not recognize the situation with only small children, ``Pedestrians stopped excluding only children'' was not labeled because that situation does not occur in Experiment II. The time during which a pedestrian stops in front of the advertiser was also annotated, instead of the paths of each scenario. This annotation was applied to all the pedestrians who passed in front of the robot in the video. To ensure valid results, video data annotation was also performed by two coders. One was the author, Y. O., and the other was a person unrelated to this study. The data for 1 day overlapped, and the analysis of the overlapped data indicates that they were well matched (Cohen's Kappa was .917).

\subsection{Results}
\begin{table}[!t]
  \renewcommand{\arraystretch}{1.3}
  \caption{Total number of pedestrians, pedestrians who stopped, and pedestrians who received the voucher.}
  \label{tab:results3}
  \centering
  \begin{tabular}{lcccc}
    \hline
     & Whole & Pedestrians & Pedestrians \\
    Human Advertisers & Pedestrians & Stopped & Received \\ \hline 
    1 (Weekday) & 1312 & 46  & 28  \\
    2 (Weekday) & 1630 & 57  & 25  \\
    3 (Weekend) & 8582 & 70  & 36  \\
    4 (Weekend) & 7103 & 290 & 205  \\
    \hline
  \end{tabular}
\end{table}
\begin{figure*}[!t]
  \centering
  \subfloat[SR]{\includegraphics[width=2.3in]{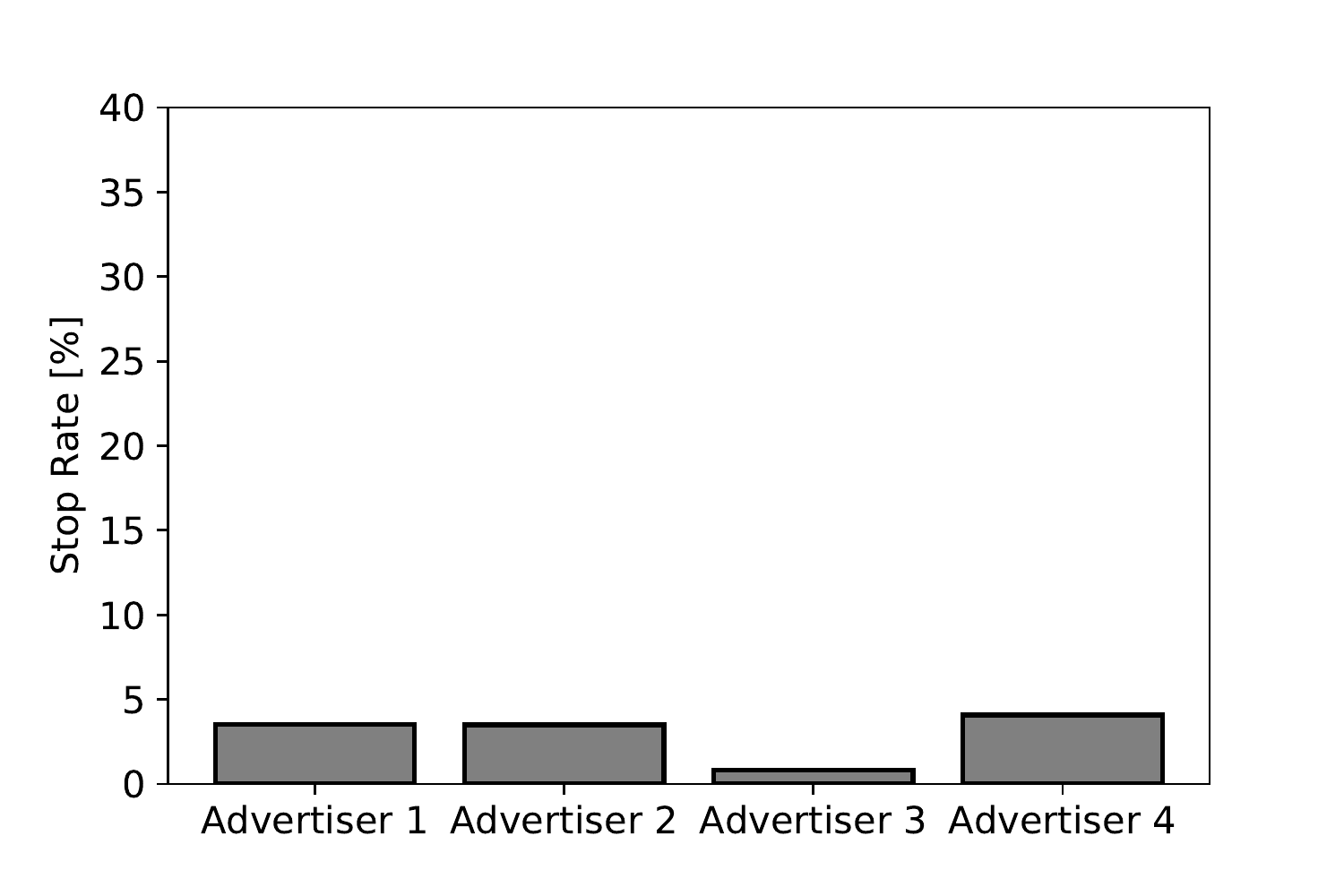}
    \label{fig:SR_human}}
  \hfil
  \subfloat[DSR]{\includegraphics[width=2.3in]{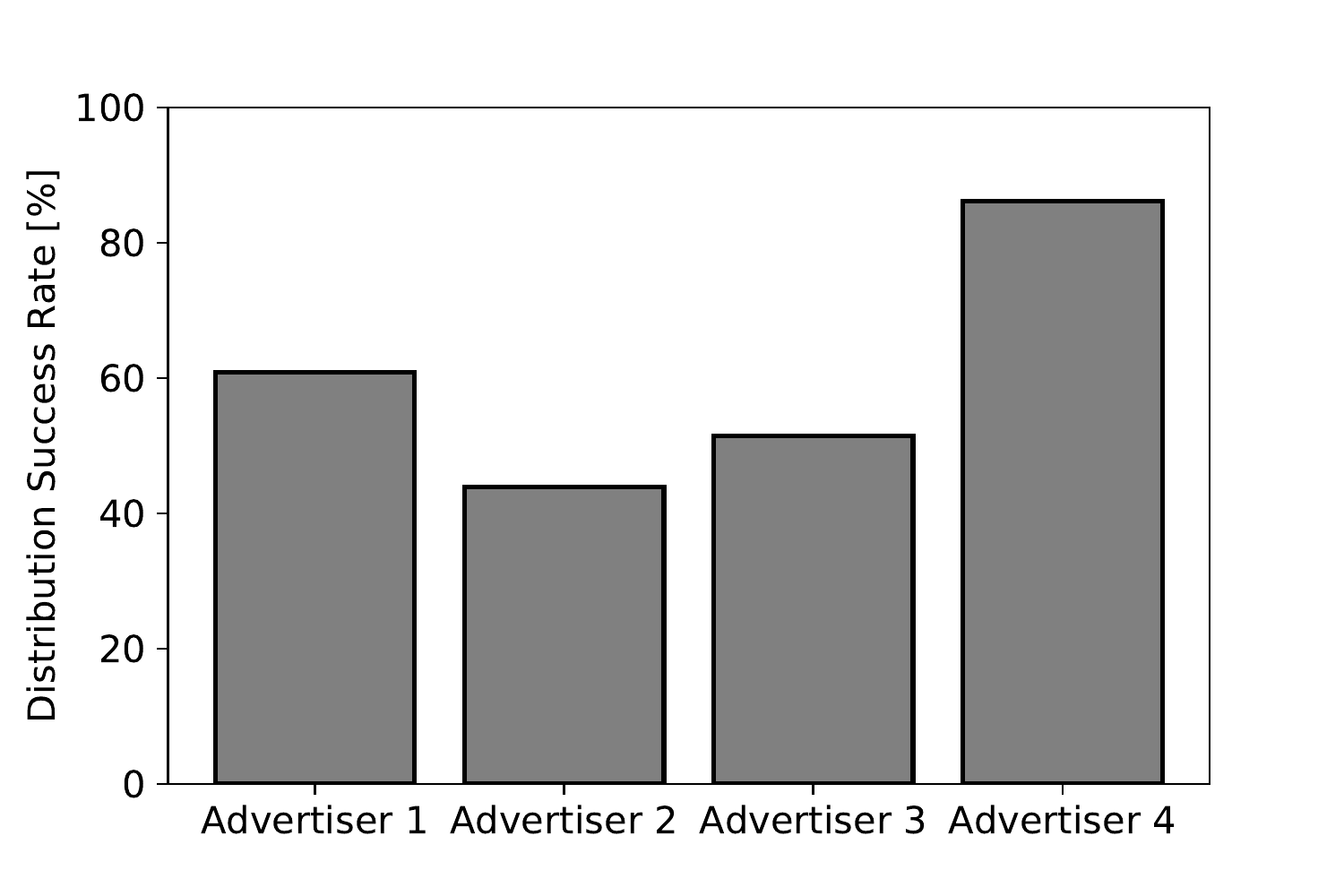}
    \label{fig:DSR_human}}
  \hfil
  \subfloat[WDSR]{\includegraphics[width=2.3in]{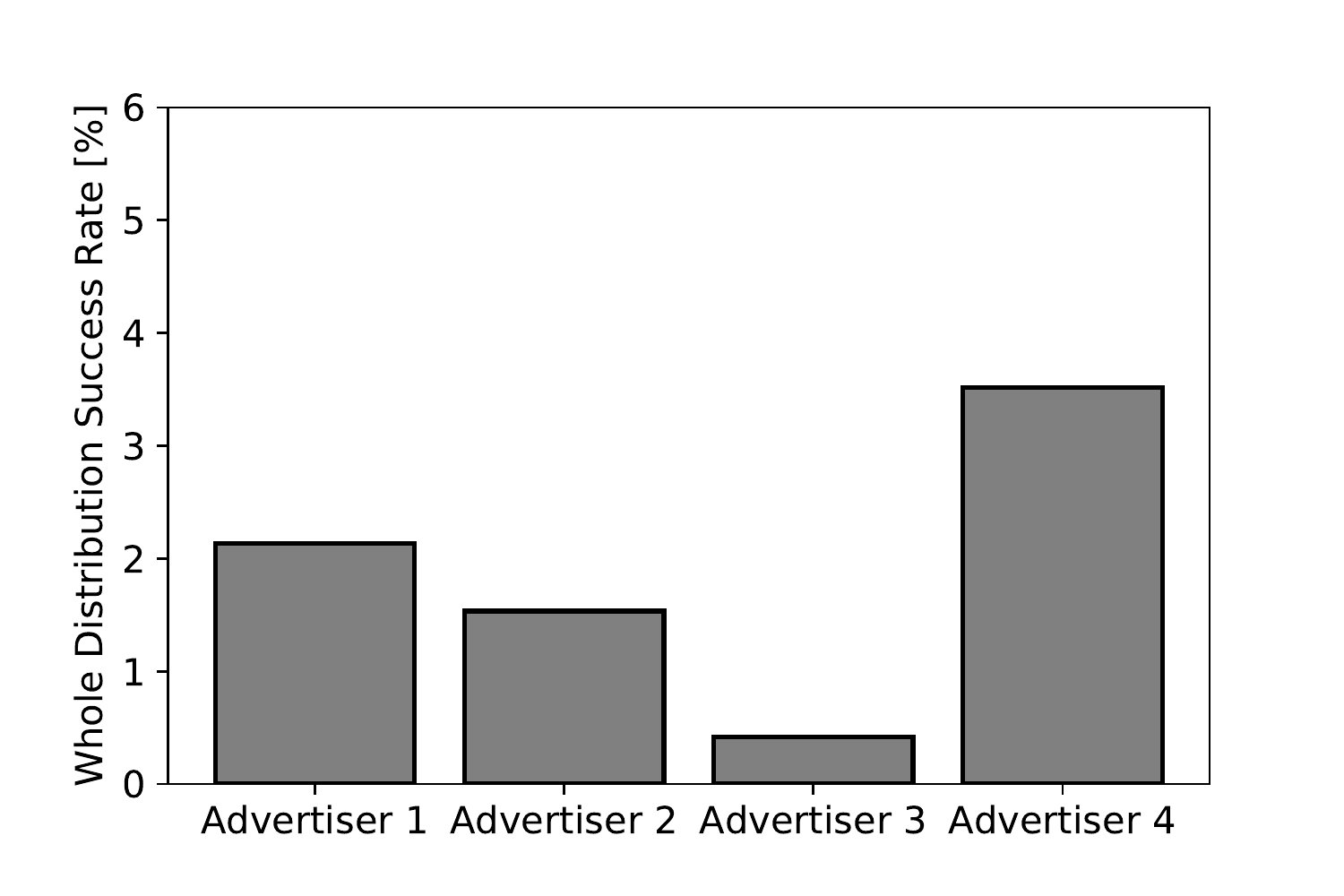}
    \label{fig:WDSR_human}}
  \caption{Results of the stop rate (SR), distribution success rate (DSR), and the whole distribution success rate (WDSR) according to each human advertiser in Experiment II.}
  \label{fig:whole_human}
\end{figure*}
\begin{figure}[!t]
  \centering
  \includegraphics[width=2.5in]{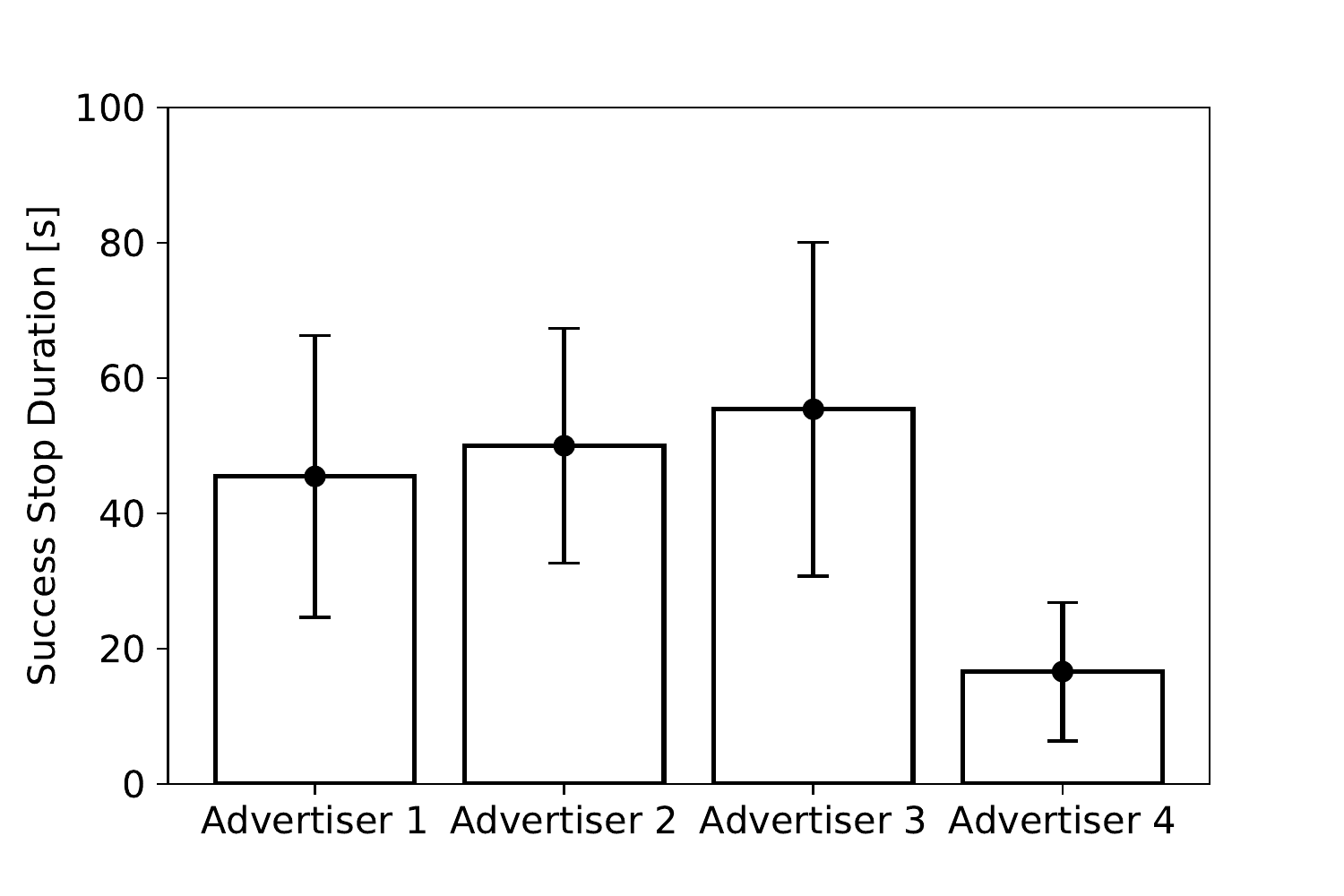}
  \caption{Results of the duration when successfully making the pedestrians stop during in Experiment II. The error bars represent the standard error of the mean.}
  \label{fig:duration_human}
\end{figure}

The results on the number of labeled pedestrians are shown in Table~\ref{tab:results3}. The results of each SR, DSR, and WDSR according to each advertiser are shown in Fig.~\ref{fig:whole_human}. The results show that there are large individual differences in all the SR, DSR, and WDSR evaluations. Advertiser 3 had the lowest SR, and Advertiser 2 had the lowest DSR. In terms of the WDSR as the total index to provide information, Advertiser 4 was able to hand out the most vouchers to the pedestrians. In an additional analysis, we measured how long the pedestrians listened to the advertiser's message when the pedestrian got the voucher, as shown in Fig.~\ref{fig:duration_human}. Advertiser 4 required the least amount of time to provide information before handing out the vouchers, with an average time of less than 20 s. These results indicate that the shorter the time to provide the information, the better the DSR.

Next, to examine the aims of Experiment II, we compared the results that were generated by the robot behaviors and that of the human advertisers. In terms of the SR, the average results of the robots for the weekday and weekend were 5.7, 10.6, and 7.7 \% for the greeting, troubling, and dancing behaviors, respectively. In contrast, the results by each advertiser were 3.5, 3.5, 0.8, and 4.1 \%, respectively. All the results in the SR by the robot outperform the human performance, and the statistical results with the Chi-squared test with the Cramer's $V$ compared the total values of the robots to the advertisers also displayed significant differences: $(\chi^2(1) = 636.84, p < 0.01, V = 0.08)$. In other words, the robots can perform the stopping task easier.

With the DSR, the average results of the robots for the weekday and weekend were 33.9, 42.0, and 26.0 \% for the greeting, troubling, and dancing behaviors, respectively. In contrast, the results by each advertiser were 60.9, 43.9, 51.4, and 86.2 \%. All the results in the DSR by the advertisers outperform the robot performance. The results of comparing the total values of the robots and the advertisers also show significant differences: $(\chi^2(1) = 158.51, p < 0.01, V = 0.19)$. This is the opposite result of the SR. We also measured the length of the stop time by randomly sampling approximately 20\% of the pedestrians who stopped in front of the robot in Experiment I. The results show that the average length of stop time for all robot behaviors was 21.5 s, whereas the average length of the stop time for all human advertisers was 22.0 s. We verified the differences in these results with a non-paired t-test, and the result showed no significant differences: $(t(678) = 0.25, p = 0.80, d = 0.02)$. This indicates that, although the length of stop time is the same for robots and human advertisers, the DSR is higher for the latter.

With the WDSR as the total index in providing the information, the average results by the robots during the weekday and weekend were 1.6, 3.6, and 1.6 \% for the greeting, troubling, and dancing behaviors, respectively. In contrast, the results by each advertiser were 2.1, 1.5, 0.4, and 3.5 \%. The results of the greeting and dancing behaviors are comparable to the average human performance. On the other hand, in terms of the troubling behavior, the robot has a higher performance than Advertiser 4, who demonstrated the best performance among the advertisers. The results in comparing the total values of the robots and the advertisers indicate significant differences: $(\chi^2(1) = 17.14, p < 0.01, V = 0.01)$. Consequently, these results show that the robots are able to perform similar to humans or even better.

\begin{table*}[!t]
  \renewcommand{\arraystretch}{1.3}
  \caption{Ratio of the number of times between one pedestrian and the group that stopped in front of the human advertiser and listened to the their message.}
  \label{tab:results4}
  \centering
  \begin{tabular}{ccccc}
    \hline
    Human & \multicolumn{2}{c}{Ratio of pedestrians stopped} & \multicolumn{2}{c}{Ratio of pedestrians received} \\
    Advertisers & One pedestrian \% & Two or more pedestrians \% & One pedestrian \% & Two or more pedestrians \% \\ \hline
    1 (Weekday) & 65.6 & 34.4 & 52.9 & 47.1 \\
    2 (Weekday) & 59.5 & 40.5 & 64.7 & 35.3 \\
    3 (Weekend) & 17.1 & 82.9 & 16.7 & 83.3 \\
    4 (Weekend) & 33.6 & 66.4 & 34.1 & 65.9 \\
    \hline
  \end{tabular}
\end{table*}

Finally, we show the ratio of the number of times between one pedestrian and the group that stopped in front of the human advertiser and listened to their message, as shown in Table~\ref{tab:results4}. The results revealed that the differences between the strategies of each human advertiser are large. Advertisers 1 and 2 had a large percentage of successful stops and distributions for one pedestrian because they likely tend to talk to a large number of individual pedestrian. Meanwhile, Advertisers 3 and 4 had a large percentage of stops and successful distributions for groups. These results in Experiment II differed from Experiment I, indicating that the results are independent of the robot behaviors.

\section{Discussions}
\label{sec5}
\subsection{Discussion on Experiment I}
This study mainly aims to investigate whether the humanoid robot can make pedestrians stop in front of it and maintain engagement with them. Thus, we designed three behaviors: greeting as an active concept, troubling as a passive-negative concept, and dancing as a passive-positive concept.

The results in Experiment I show that when the robot exhibited the troubling behavior, it had the best performance among the proposed behaviors in terms of attracting pedestrians and providing information. In addition, the dancing behavior also showed a similar high performance in comparison to troubling only to make people stop. The difference in these results is probably due to the different reasons why the pedestrians approached the robots. When humans see the weak or human-dependent robots, they tend to increase their engagement with the robots because humans want to help robots~\cite{Khaoula14}. This is similar to a phenomenon wherein an adult reaches out to a child when the child is in trouble. Therefore, it seems that many pedestrians were willing to listen to the robot in the troubling situation because the troubling behavior is a similar case. In contrast, in the dancing behavior, we assume that the most common reason that pedestrians approached the robot is for fun. However, despite that they approached the robot to enjoy its dance, the pedestrians felt disconnected when the robot started talking about the shop's information after they approached it. Thus, it can be assumed that the DSR was the lowest among all the proposed behaviors owing to this gap. Consequently, these results suggest that a passive concept (troubling and dancing) is effective to make the pedestrians stop, when compared to the active concept (greeting). In addition, in the passive concept, the consistency in the robot's behavior is crucial in maintaining its engagement with the pedestrians.

Other interesting results are the gender and age differences in the SR and DSR. The results of the SR and DSR showed that women, as well as children, mostly stop in front of the robot and listen when the robot talks. The same situation has also occurred in other studies~\cite{Bergstrom09}, wherein children, sometimes accompanied by their parents, often interact with the robots. In addition, the results of the previous studies indicated that men have a more positive attitude toward interacting with robots than women~\cite{Mutlu06, Nomura17}. However, other studies that use the robot ``Sota,'' which is the same as used in the present study,~\cite{Okafuji20b} showed that women are more interested in the robot; this finding is consistent with our results. Therefore, the results suggest that gender difference in terms of the interest in the robots does not give a definite decision, but it may depend on the appearance of the robot. Meanwhile, the results of the DSR did not show a significant difference by the age. This robot was a non-interactive robot; that is, it did not have the ability to communicate with humans through dialogues. In this case, we assume that the results of the DSR represent the pedestrian's time where they lost interest in the robot and it was similar regardless of the age. In other words, even if the robots are interacting with a person who is strongly interested in the robot, a robot with a poor interactive ability will quickly get boring.

Finally, it was determined that, when pedestrians are in a group, they are more likely to stop and are more willing to listen to the full message. Several examples of these observations have been reported in other studies~\cite{Shi18, Michalowski06, Tonkin17, Okafuji20b}. These observations were thought to be caused by another person who draws the pedestrian to interact with the robot, and not by the robot itself~\cite{Shi18, Michalowski06}. Therefore, in this study, the reason for this observation can be assumed to be that if one person in the group is interested in the robot, the others should listen to the robot; thus, they will wait for him and/or her. Therefore, in crowded situations, such as shopping malls, it is more efficient if the robots talk to the group to provide the information than to an individual person.

\subsection{Discussion on Experiment II}
We also compared the results by the robots and the humans in Experiment II. When comparing both their performance, the SR was higher for the robots and the DSR was higher for the humans. The reason for the low DSR can be assumed to be that the robot's verbal interaction ability was significantly lower than that of humans. It was not possible to implement verbal interaction in the robotic system because the robots do not exhibit a proper speech recognition in a noisy commercial environment. In future studies, this problem can be solved by developing a technology that can recognize the speech of pedestrians correctly even in noisy environments.

In contrast, there are several possible reasons why, from the SR perspective, the robots can perform better than humans. The first is simply because pedestrians were strongly interested in the robot's behavior. The greeting behavior, which is the lowest SR among the robot behaviors, is able to improve the SR in comparison to humans owing to its novelty effects~\cite{Gockley05}. In addition, other behaviors were able to improve the SR drastically. Thus, the different types of robot behaviors can attract pedestrians. The second was that the pedestrians did not prefer the situation wherein the adult's human advertisers were calling out to them to provide information. As mentioned in the discussion on Experiment I, an adult tends to reach out to a child when the child is in trouble. On the other hand, in the case of Japan, where the experiment was conducted, when a strange adult talks to a person in a shopping mall or in a town, the stranger is often a salesperson who recommends something. This is considered nuisance to the pedestrians. Therefore, pedestrians avoid conversations with strangers before they know the type of information will be provided. Accordingly, we believe that the SR of the adult advertisers was lower; however the SR could be higher if a child advertiser did the same task (if it were ethically possible).

These results may be useful in the future collaborative design of robots to support humans. For example, by having a remote avatar robot system such as in~\cite{Baba20, Shiomi08}, it is possible to build a more high-performance system by integrating the capabilities of robots with those of humans. In this study, the environment was a noisy shopping mall in which it is difficult to recognize the speech of a particular individual. Thus, we constructed a passive medium system in which the robot talks unilaterally. However, when it comes to interacting with pedestrians, humans can interpolate the dialogue to improve the interactivity. The autonomous robot attempts to make pedestrians stop in front of the robot, and humans are interpolated in situations where it is difficult for the autonomous robot to talk with the pedestrians. Thus, we expect to build a system that can demonstrate a high performance by interpolating the weaknesses of the robots and humans.

The results of the WDSR for the greeting and dancing behaviors of the robots are comparable to human performance; however, the troubling behavior performance exceeded all the results of the human advertisers. While the SR and DSR are part of the process in evaluating the performance of an informational task, the WDSR is the final evaluation of performance. In other words, in this experiment, the robot succeeded in providing more information to the pedestrians than the human advertisers. Therefore, the results may indicate that robots are more effective in the information provision task.

In addition, this result does not consider the decrease in human performance over time. In this experiment, the advertisers were asked to perform the tasks for 3 h, and performance degradation during this time was not measured. However, if they perform the task over a longer period, performance degradation due to fatigue occurs. In that case, the robots can deliver better results than humans in the work environment. 

In Experiment II, we observed that the SR and WDSR of Advertiser 3 were lower than those of the other advertisers. This is because human advertiser 3 approached pedestrians less often than the other advertisers (Number of people passing per approach for each advertiser: 4.6, 5.6, 46.9, and 10.9, respectively). On the other hand, this phenomenon does not occur with robots; thus, robots have the advantage in that their performance is not affected by factors such as human individual differences. In summary, the results from this study suggest that robots can be sufficient as a labor support technology, which is one of the goals of robotic research.

\subsection{Limitations}
Finally, we want to present the limitations of this study. First, we compared the performances of the three types of proposed behavioral concepts of robots. This is because we considered that the behavioral concepts can be used generically, which would be more effective in comparison to detailed robot behavior. However, it is unclear whether two types of robots that are implemented with the same behavioral concept but have slightly different details of motion can achieve similar results. Even when we use the passive method, we can design other behaviors than dancing. This is a limitation that we need to explore in future studies, which also considers the difference in the robot's appearance and degrees of freedom. 

In addition, this study cannot demonstrate whether the proposed behavioral concept always shows the same results. In this study, we conducted the experiment with the robot in a commercial facility where many people have relatively more time to spare. However, through the experiments, we found that most of the approaches from the robot failed for people in a hurry. Therefore, depending on environmental conditions such as the context and location where the robot is installed, we cannot guarantee if the troubling behavior has as a significant effect on people as the results of this study.

Next, our results strongly depend on the novelty effect. In the troubling behavior, the robot said ``I'm in trouble'' to attract pedestrians' attention, and then they conveyed information about the store. This may have been a type of ``crying wolf.'' Further, in the field of HRI, for example, it has been reported that robot errors decrease people's trust in the robot~\cite{Salem15}. In other words, when people are exposed to the troubling behavior more than once, their trust in the robot may decrease, and they may not listen to the robot. In such a situation, the robot may not be able to surpass human performance even with SR that have shown superior performance over humans. This long-term performance will be considered for future studies.

We should also consider that the results of this study are highly dependent on cultural differences. Previous studies on human-robot interaction with several cultural differences have shown that people with different cultures behave differently depending on the task and the appearance of the robot~\cite{Li10, Trovato13}. For example, for people who think robots are mechanical rather than humans, a robot's troubling behavior may seem creepy. In this case, troubling behavior could deliver the worst result. In human--human interaction, we showed that pedestrians in Japan tend to avoid talking with strangers. However, in other cultures, robots may not be able to outperform humans in terms of the SR results. During this experiment, we did not interview any pedestrians who interacted with the robot. As we cannot infer how the pedestrians felt through their interaction with the robot, we cannot have a rigorous discussion on the cultural differences that affected them. Therefore, the cultural difference is another limitation of this study.

From a perspective closer to cultural differences, the degree to which people are accustomed to robots affects the results of this study. In today's society, social robots are still a rarity and an intriguing object. However, we believe that in a future society where a variety of social robots are prevalent, people care less about what a robot does even if it dances or behaves as it is in trouble. Therefore, we need to consider that, in the future, results may be different from those obtained in this study.

\section{Conclusion}
\label{sec6}
This study investigated whether a humanoid robot can make pedestrians stop in front of it and listen to its message. We proposed three types of behavioral concepts for the robots for the general purpose: active, passive-negative, and passive-positive concepts. According to the three proposed concepts, we specifically designed the ``greeting behavior'' as active, ``troubling behavior'' as passive-negative, and ``dancing behaviors'' as passive-positive. The robot with each behavior was placed in a shopping mall, and the effectiveness of the robots for providing information was verified.

The results from the exploratory field experiments revealed that the troubling behavior, that is, the robot behaves as if it is in trouble, can make pedestrians stop more and stay longer in front of the robot. These results were compared to the results achieved by four humans under the same situation, in which they attempted to make the pedestrians stop to provide information. The comparative results show (1) the performance of the robots were higher than that of the humans in the stop rate (SR), and (2) in the distribution success rate (DSR), the human performance was better than the robots' performance. In particular, in terms of the whole distribution success rate (WDSR), the performance obtained using the greeting and dancing behaviors of the robots are comparable to the human performance. Furthermore, it was determined that the performance of the troubling behavior was higher than those of all the human advertisers who participated in this experiment. These findings demonstrate that the performance of robots is not inferior to that of humans in providing information task. Therefore, it is expected that service robots are able to perform well in the real world. In other words, the results of this study suggest that robots can be sufficient as a labor support technology, which is one of the goals of robotic research.

This study, however, has some limitations because it is difficult for the robots to interact naturally with pedestrians in a noisy environment. This is because automatic dialogue generation is difficult due to the low accuracy of speech recognition for certain pedestrians in noisy environments. These problems are common to all robots that operate in real environments. Hence, ensuring that the robot can recognize the speech content of only the target person in a noisy environment is required. However, by interpolating the weaknesses of the robots and humans, we can build an integrated robot system that can demonstrate high performance. By achieving this, we believe that it is important for robot designs to compensate for the weaknesses of robots and humans in the future.

%
\IEEEpeerreviewmaketitle

\ifCLASSOPTIONcaptionsoff
  \newpage
\fi

\end{document}